\documentclass{article}
\usepackage{ijcai26}

\usepackage{times}
\usepackage{url}
\usepackage[hidelinks]{hyperref}
\usepackage[utf8]{inputenc}
\usepackage[small]{caption}
\usepackage{graphicx}
\usepackage{amsmath}
\usepackage{amssymb}
\usepackage{amsthm}
\usepackage{booktabs}
\usepackage{algorithm}
\usepackage{algorithmic}
\usepackage[switch]{lineno}
\usepackage{multirow}
\usepackage{xspace}
\usepackage{subcaption}
\usepackage{xcolor}

\newcommand{\eg}{\textit{e.g.}}
\newcommand{\ie}{\textit{i.e.}}

\title{Erase at the Core: Representation Unlearning for Machine Unlearning}

\author{
Jaewon Lee,
Yongwoo Kim,
Donghyun Kim\footnote{Corresponding author.}\\
\affiliations
Korea University\\
\emails
jwlee010222@gmail.com, \{yongwookim, d\_kim\}@korea.ac.kr
}
\begin{document}

\maketitle

\begin{abstract}
Many approximate machine unlearning methods demonstrate strong logit-level forgetting---such as near-zero accuracy on the forget set---yet continue to preserve substantial information within their internal feature representations. We refer to this discrepancy as superficial forgetting. Recent studies indicate that most existing unlearning approaches primarily alter the final classifier, leaving intermediate representations largely unchanged and highly similar to those of the original model. To address this limitation, we introduce the Erase at the Core (EC), a framework designed to enforce forgetting throughout the entire network hierarchy. EC integrates multi-layer contrastive unlearning on the forget set with retain set preservation through deeply supervised learning. Concretely, EC attaches auxiliary modules to intermediate layers and applies both contrastive unlearning and cross-entropy losses at each supervision point, with layer-wise weighted losses. Experimental results show that EC not only achieves effective logit-level forgetting, but also substantially reduces representational similarity to the original model across intermediate layers. Furthermore, EC is model-agnostic and can be incorporated as a plug-in module into existing unlearning methods, improving representation-level forgetting while maintaining performance on the retain set.

\end{abstract}

\section{Introduction}

Recent data protection regulations, most notably the EU General Data Protection Regulation (GDPR), formalize a ``right to be forgotten,'' under which individuals can request the erasure of personal data and the withdrawal of its influence from deployed machine learning systems. Fulfilling such requests in practice requires not only deleting raw records but also removing their impact on trained models. Machine unlearning (MU) addresses this requirement by aiming to remove the influence of a designated subset of training data from a deployed model, typically termed the \emph{forget set}, while preserving performance on the remaining \emph{retain set}~\cite{cao2015towards,bourtoule2021machine}. This capability is important both for regulatory compliance and for correcting models trained on corrupted or personal data. The most direct way to implement MU is to retrain the model from scratch on the retain set only, which guarantees exact removal of the forget data but is prohibitively expensive for modern large-scale deep neural networks~\cite{bourtoule2021machine,aldaghri2021coded}. As a result, recent research has focused on \emph{approximate} unlearning algorithms that update a pretrained model to emulate the behavior of the retrained model at a fraction of the computational cost.

Approximate MU methods span a wide design space, ranging from loss-based approaches such as gradient ascent and random relabeling~\cite{golatkar2020eternal,thudi2022unrolling}, to distillation-based and regularization-based strategies~\cite{kurmanji2023towards,zhou2025decoupled,chen2024unsc}, and more recently to representation-based methods that leverage metric learning or contrastive objectives~\cite{cotogni2023duck,zhang2024contrastive}. These methods are typically evaluated using logit-based metrics, including forget set accuracy, retain set accuracy, and membership inference attack (MIA) success rates~\cite{shokri2017membership,carlini2022membership}.

However, recent studies~\cite{kim2025arewe,jeon2024idi,siddiqui2025dormant} reveal that achieving near-zero forget accuracy and low MIA success rates does not guarantee complete removal of forget set information. Representation-based analyses using Centered Kernel Alignment (CKA)~\cite{kornblith2019similarity} and the Information Difference Index (IDI)~\cite{jeon2024idi} demonstrate that many unlearning algorithms leave intermediate features highly similar to the original model, indicating persistent \emph{feature residuals} despite seemingly successful logit-level forgetting~\cite{kim2025arewe,jeon2024idi}. Moreover, linear probing attacks, which freeze the backbone and retrain only a final classifier, can recover substantial forget set accuracy from supposedly unlearned models~\cite{jeon2024idi,jung2025opc}. We refer to this phenomenon as \emph{superficial forgetting}: the model achieves forgetting only in the final classifier, while retaining linearly separable structure for forget classes in its intermediate representations. 
These findings call for unlearning mechanisms that operate throughout the depth of the network, not just at the output layer. Although representation-based unlearning methods such as DUCK~\cite{cotogni2023duck} and Contrastive Unlearning (CU)~\cite{zhang2024contrastive} manipulate embeddings in a feature space, recent evaluations show that even these approaches can leave non-negligible similarity to the original model in intermediate layers. This motivates an approach that explicitly enforces erasure at the core by pushing the representations of forget samples away from their original embeddings across multiple layers.


To address this, we propose \emph{Erase at the Core} (EC), a representation-based unlearning framework designed to eliminate remaining knowledge from shallow to deep layers. We apply unlearning objectives to the forget set while simultaneously enforcing knowledge-preservation losses on the retain set across layers.  Motivated by prior work on deep supervision~\cite{lee2015dsn,zhang2022contrastivedeepsup}, we start from the original model (\eg, ResNet-50~\cite{he2016resnet}) and initialize auxiliary modules at intermediate layers using supervised contrastive learning~\cite{khosla2020supervised}. During unlearning, we extend the contrastive unlearning objective of CU~\cite{zhang2024contrastive} from a single-layer setting to a multi-layer regime: the forget embeddings at each supervised layer are diffused into the manifold of retain samples, while cross-entropy losses on retain samples maintain classification utility. By assigning progressively larger weights to deeper layers---where high-level, class-discriminative features are encoded---this deep supervision ensures that forgetting signals propagate through the entire feature hierarchy.


In addition, we revisit existing strong unlearning baselines (\eg, \cite{chen2024unsc,cotogni2023duck,bonato2024retain,kurmanji2023towards,golatkar2020eternal,fan2024salun,zhou2025decoupled,zhang2024contrastive}) with both logit-based and representation-based evaluation~\cite{kim2025arewe,jeon2024idi} metrics from prior work, and provide a unified and comprehensive evaluation across diverse settings. Specifically, we conduct a comprehensive evaluation on a large-scale multi-class unlearning scenario based on ImageNet-1K~\cite{deng2009imagenet}~\cite{kim2025arewe,kornblith2019similarity}, where 100 classes are designated as the forget set and the remaining 900 as the retain set. Beyond standard logit-based metrics on forget and retain sets, we employ representation-based evaluation using layer-wise CKA~\cite{kim2025arewe} and IDI~\cite{jeon2024idi} over the final layer blocks on the forget set. Our experiments show that EC consistently outperforms existing unlearning baselines across diverse evaluation metrics and achieves substantially greater divergence from the original model. We further investigate how EC can be serve as a plug-in module for other unlearning algorithms.


Our contributions are summarized as follows:
\begin{itemize}
    \item We introduce EC (Erase at the Core), a multi-layer unlearning framework that combines contrastive unlearning with deep supervision to enforce core feature forgetting while preserving retain set utility.
    \item We revisit existing unlearning baselines with comprehensive evaluation using logit-based and representation-based metrics (CKA, IDI, and k-NN downstream task performance) on large-scale multi-class unlearning. EC moves the learned representations farther from the original model than prior methods.
    \item We conduct extensive experiments across benchmarks (ImageNet-1K, CIFAR-100), forgetting scenarios (random classes and top-similarity classes defined with respect to a downstream dataset), and architectures (ResNet-50, Swin-Tiny), demonstrating the robustness and effectiveness of EC across diverse settings.
    \item We show that EC is model-agnostic and can be applied as a plug-in module to other representation-based unlearning methods, improving their forgetting strength at the representation level.
\end{itemize}

Together, these results highlight the importance of moving from superficial, logit-based forgetting toward deep (core), representation-based forgetting, and position EC as a practical step toward stronger unlearning in large-scale unlearning scenarios.

\begin{figure*}[t]
    \centering
    \includegraphics[width=0.90\linewidth]{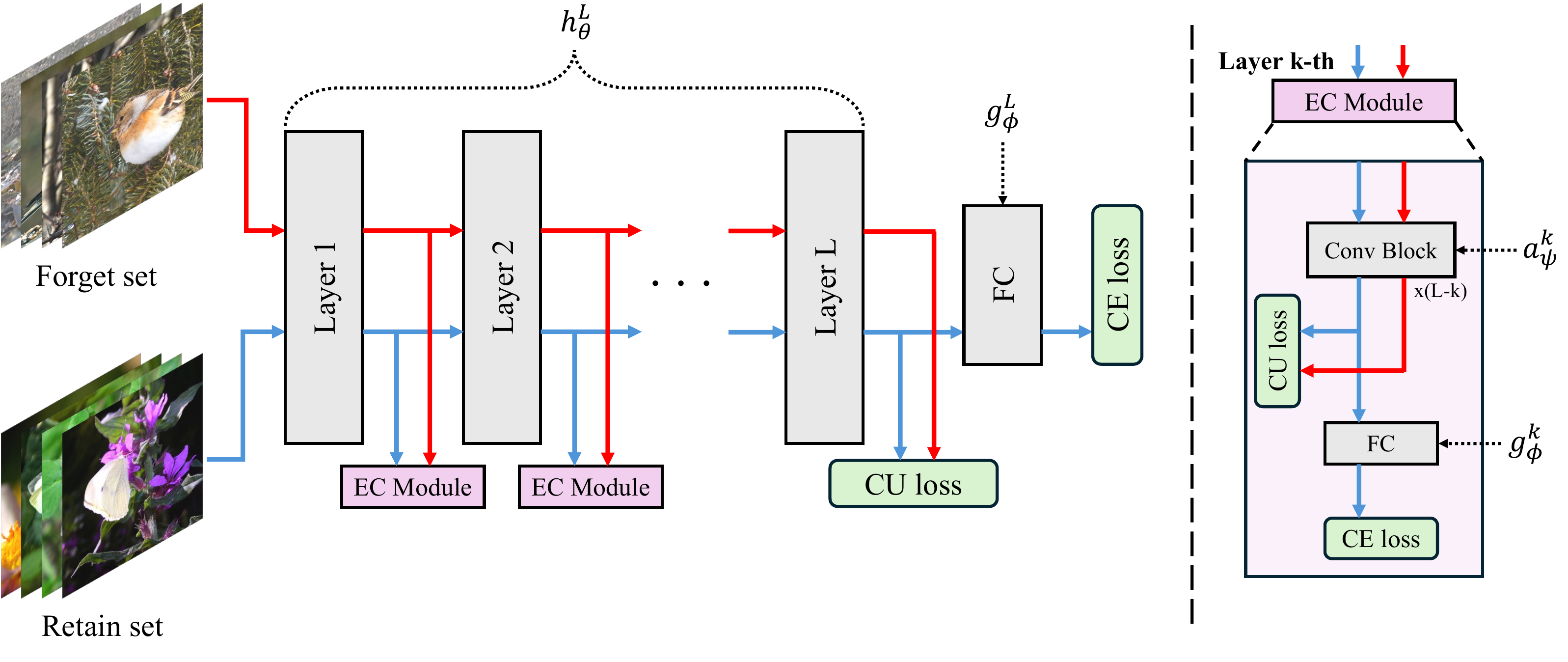}
    \caption{Illustration of Erase at the Core (EC). EC attaches EC Modules to intermediate layers and applies layer-wise contrastive unlearning loss along with the cross-entropy loss. Here, $L$ denotes the number of layers in the backbone, and the EC module attached after the $k$-th layer consists of $(L-k)$ Conv Blocks.}
    \label{fig:overall architecture}
\end{figure*}

\section{Related Work}
\label{sec:relatedworks}

\subsection{Machine Unlearning}
Machine unlearning aims to selectively remove the influence of specific data subsets (\ie, the forget set) from a trained model while preserving the performance on the remaining data (\ie, the retain set). This field is primarily categorized into exact and approximate unlearning. Exact unlearning methods (\eg, SISA~\cite{bourtoule2021machine}, ARCANE~\cite{yan2022arcane}) theoretically guarantee complete data removal but incur prohibitive computational costs for large-scale deep neural networks. Early approximate methods focused on loss inversion: Gradient Ascent (GA)~\cite{golatkar2020eternal,thudi2022unrolling} maximizes forget set loss, while Random Labeling (RL)~\cite{golatkar2020eternal} replaces forget labels with random ones to disrupt learned associations. To better preserve retain set utility, distillation-based approaches were introduced. SCRUB~\cite{kurmanji2023towards} formulates unlearning as a teacher-student objective that simultaneously pulls the model toward the original on retain data and pushes it away on forget data.


Recent unlearning baselines operate on feature spaces. DUCK~\cite{cotogni2023duck} and Contrastive Unlearning~\cite{zhang2024contrastive} apply metric learning and contrastive objectives to push forget samples away from their class neighborhoods in the embedding space. Other methods (PL~\cite{chen2024unsc}, SCAR~\cite{bonato2024retain}, DELETE~\cite{zhou2025decoupled}) employ null-space projection, Mahalanobis-distance criteria, or mask distillation to balance forgetting and retention. Despite these advances, most methods primarily operate on the last classifier or logits, leaving intermediate layers vulnerable to information leakage.

\subsection{Unlearning Evaluation: Logit-based and Representation-based Evaluation}

A critical limitation of many approximate unlearning methods is the phenomenon of ``superficial forgetting,'' where models suppress output logits for the forget set while retaining significant information in internal feature representations. Recent studies reveal that low forget set accuracy does not necessarily imply complete erasure of underlying knowledge. \citeauthor{kim2025arewe}~\shortcite{kim2025arewe} demonstrate that despite achieving near-zero forget accuracy, existing baselines maintain feature representations highly similar to the original model, as measured by Centered Kernel Alignment (CKA)~\cite{kornblith2019similarity}, arguing that true unlearning requires features to diverge from the original model. Similarly, \citeauthor{jeon2024idi}~\shortcite{jeon2024idi} show that intermediate layers often retain high Mutual Information (MI) with forget labels even when outputs appear random, and propose the Information Difference Index (IDI) to quantify residual information in intermediate layers by comparing the unlearned model to a retrained baseline.

The vulnerability of superficial forgetting is further exposed by linear probing attacks. \citeauthor{jung2025opc}~\shortcite{jung2025opc} and \citeauthor{jeon2024idi}~\shortcite{jeon2024idi} demonstrate that by freezing the backbone of an ``unlearned'' model and retraining only a final classifier, forget set accuracy can be easily recovered, revealing that forget features remain linearly separable. Collectively, these representation-based metrics and attacks expose a critical limitation of existing unlearning methods: the lack of mechanisms to enforce forgetting across intermediate layers.

\subsection{Representation Learning Across Intermediate Layers}

Learning intermediate representation was originally introduced to mitigate vanishing gradients and enhance feature discriminability by directly supervising intermediate layers of a network~\cite{lee2015dsn,szegedy2015going}. Building on this idea, Contrastive Deep Supervision (CDS)~\cite{zhang2022contrastivedeepsup} extends deep supervision by applying contrastive objectives at multiple layers, encouraging semantically consistent representations across depths while preventing feature collapse. In this paper, we explicitly control intermediate representations to eliminate residual features that persist in intermediate layers despite successful logit-level forgetting. We propose the Erase at the Core (EC), a framework that enforces feature-space divergence throughout the network depth by applying layer-wise supervision, ensuring that forgetting is achieved consistently across all layers of the model.

\section{Method: Erase at the Core}
As discussed in Section~\ref{sec:relatedworks}, existing unlearning methods often achieve strong logit-level forgetting yet fail to alter internal representations when assessed through metrics such as CKA~\cite{kim2025arewe}. To address this limitation, we propose EC (Erase at the Core), which enforces representation-level forgetting by applying supervision to multiple intermediate layers.

As illustrated in Figure~\ref{fig:overall architecture}, we extend the model architecture by attaching EC modules to intermediate layers of the backbone. This design enables the application of supervision not only at the final layer but also at intermediate layers throughout the network. At each supervision point, we apply two complementary objectives: (1) a contrastive unlearning loss $\mathcal{L}_{\text{CU}}^l$ that pulls forget features toward the manifold of retain set samples, thereby erasing class-specific information, and (2) a cross-entropy loss $\mathcal{L}_{\text{CE}}^l$ on retain data to preserve classification performance. Leveraging the hierarchical nature of CNNs---where earlier layers capture low-level features and deeper layers encode high-level, class-discriminative features~\cite{zeiler2014visualizing,yosinski2014transferable}---we assign progressively larger weights to deeper layers in the total loss formulation.

\subsection{Preliminaries and Problem Definition}
\label{subsec:preliminaries}

\textbf{Problem Definition.}
We denote the entire dataset by $\mathcal{D} = \{(x_i, y_i)\}_{i=1}^{N}$, where $x_i$ is an input image, $y_i$ is its corresponding label, and $N$ is the total number of samples. A subset of $\mathcal{D}$ that is requested to be forgotten is referred to as the \emph{forget set} and is denoted by $\mathcal{D}_f \subset \mathcal{D}$. The \emph{retain set} $\mathcal{D}_r$ is defined as the subset containing all remaining samples that are not in the forget set, \ie, $\mathcal{D}_r = \mathcal{D} \setminus \mathcal{D}_f$. We additionally consider disjoint test sets that share the same classes as the training sets, denoting the test forget set and test retain set as $\mathcal{D}_f^{\text{te}}$ and $\mathcal{D}_r^{\text{te}}$, respectively.

Let $f_o(\cdot)$ be the \emph{original model} pretrained on the full dataset $\mathcal{D}$. The goal of machine unlearning is to construct an \emph{unlearned model} $f_u(\cdot)$ that effectively removes the influence of the forget set $\mathcal{D}_f$ from $f_o(\cdot)$ while preserving the influence of the retain set $\mathcal{D}_r$. As a gold-standard reference, we denote by $f_r(\cdot)$ the \emph{retrained model} obtained by training from scratch using only the retain set $\mathcal{D}_r$.

We consider a classification model composed of a backbone feature extractor $h_\theta(\cdot)$ and a final classifier $g_\phi(\cdot)$, such that $f(x) = g_\phi(h_\theta(x))$.
We denote by $z = \mathrm{Norm}\big(h_\theta(x)\big)$ the $L_2$-normalized backbone feature (penultimate-layer representation), which is used as the embedding for 
classification. 

\noindent\textbf{Multi-class Forgetting Scenarios.}
Machine unlearning scenarios can be categorized by how the forget set is specified: sample-wise forgetting, where individual data points are removed, and class-wise forgetting, where all samples belonging to specific classes are removed. In this work, we focus on multi-class forgetting, a large-scale class-wise unlearning setting where the forget set $D_f$ contains all samples from a designated subset of classes $C_f$, and the retain set $D_r$ contains samples from the remaining classes $C_r$, with $C_f \cap Cr = \emptyset$.

\subsection{Architectures}
\label{subsec:architecture}
We attach EC modules after intermediate layers (stages) of the backbone. This design is architecture-agnostic and can be applied to diverse backbones such as ResNet-50~\cite{he2016resnet} or Swin-Tiny~\cite{liu2021swin}. We choose ResNet-50 as the default backbone. 
The Conv Block within each EC module follows the architecture proposed in Contrastive Deep Supervision (CDS)~\cite{zhang2022contrastivedeepsup}; detailed specifications are provided in the Technical Appendix.

Each Convolutional (Conv) Block is initially trained using Supervised Contrastive Learning (SupCon)~\cite{khosla2020supervised} on the entire dataset $\mathcal{D}$, while keeping the backbone and the final classifier frozen. The classifiers attached after each Conv Block are randomly initialized and remain untrained during this phase; they are learned through the cross-entropy loss during the unlearning stage. Throughout the unlearning process, the backbone, final classifier, and all EC modules are jointly updated according to their respective objectives.

\begin{algorithm}[tb]
    \caption{EC: Erase at the Core}
    \label{alg:EC}
    \textbf{Input}: Backbone $h_{\theta}$ and final classifier $g_{\phi}$, EC modules without FC $a_\psi$,
    Forget set $\mathcal{D}_f$, Retain set $\mathcal{D}_r$, number of layers $L$,
    hyperparameters $\lambda_{\text{CU}}, \lambda_{\text{CE}}, \tau, w_{1-L}$\\
    \textbf{Output}: Unlearned model $(h_{\theta}, g_{\phi})$
    \begin{algorithmic}[1]
        \WHILE{not converged}
            \STATE Sample input batch $B_f$ and labels $Y_f$ from $\mathcal{D}_f$
            \STATE Sample input batch $B_r$ and labels $Y_r$ from $\mathcal{D}_r$
            \FOR{$l=1$ \textbf{to} $L$}
                \STATE $Z_f^l = \text{Norm}(a_\psi^l(h_\theta^l(B_f)))$
                \STATE $Z_r^l = \text{Norm}(a_\psi^l(h_\theta^l(B_r)))$
                \STATE $\hat{Y}_r^l = g^l_\phi(a_\psi^l(h_\theta^l(B_r)))$
                \STATE $\mathcal{L}_{\text{CE}}^l = \text{CE}(\hat{Y}_r^l, Y_r)$
                \STATE $\mathcal{L}_{\text{CU}}^l = - \frac{1}{|B_f||B_r|}\sum_{i,j}\log \text{sim}(z_i^l, z_j^l; \tau)$
                \STATE \quad where $z_i^l \in Z_f^l, z_j^l \in Z_r^l$
            \ENDFOR
            \STATE $\mathcal{L}_{\text{total}} = \sum_{l=1}^{L} w_l ( \lambda_{\text{CU}} \mathcal{L}_{\text{CU}}^l + \lambda_{\text{CE}} \mathcal{L}_{\text{CE}}^l )$
            \STATE Update $\theta, \phi, \psi$ using $\nabla \mathcal{L}_{\text{total}}$
        \ENDWHILE
        \STATE \textbf{return} updated $h_{\theta}$ and $g_{\phi}$ 
    \end{algorithmic}
\end{algorithm}

\subsection{Unlearning Objectives}
\label{subsec:unlearning_objective}

We follow the notation in Sec.~\ref{subsec:preliminaries}. 
Let $L$ denote the number of layers (stage) (\ie, $L=4$ for ResNet-50).
For each $l\in\{1,\dots,L\}$, $h_\theta^l(\cdot)$ denotes the backbone up to layer $l$, 
$a_\psi^l(\cdot)$ the auxiliary module(EC module) without FC, and $g_\phi^l(\cdot)$ the classifier after $a_\psi^l$.
We set $a_\psi^L$ to the identity mapping, so that $g_\phi^L$ coincides with the original final classifier.

\noindent\textbf{Contrastive Unlearning Loss on the Forget Set.}
We extend contrastive unlearning loss to multiple layers, applying it to our multi-class forgetting setting defined in Section~\ref{subsec:preliminaries}.

The core idea is to diffuse the forget sample embeddings into the embedding space of the retain samples, thereby erasing class-specific information associated with the forget classes. For a given layer $l$ and input $x$, let $h_\theta^l(x)$ denote the raw output feature of layer $l$. We denote our attached EC module as $a_\psi^l(\cdot)$. Accordingly, the $L_2$-normalized embedding used for the contrastive unlearning loss at each layer is defined as $z^l = \text{Norm}(a_\psi^l(h_\theta^l(x)))$.

We maximize the similarity between the embeddings of forget and retain samples:
\begin{equation}
    \mathcal{L}_{\text{CU}}^l = - \frac{1}{|\mathcal{D}_f|} \sum_{x_i \in \mathcal{D}_f} 
    \frac{1}{|\mathcal{D}_r|} \sum_{x_j \in \mathcal{D}_r} 
    \log \text{sim}(z_i^l, z_j^l; \tau),
\end{equation}
where $\text{sim}(u, v; \tau) = \exp(u \cdot v / \tau)$ denotes the temperature-scaled cosine similarity. By applying this loss across multiple intermediate layers, we enforce forgetting at the representation level throughout the network, rather than achieving only superficial forgetting at the output layer.

\noindent\textbf{Cross-Entropy Loss on the Retain Set.}
Similarly, we apply the cross-entropy loss across multiple layers. The cross-entropy loss is computed solely on the retain set $\mathcal{D}_r$. This not only maintains the model's performance on the retain set during unlearning but also contributes to more effective forgetting. The layer-wise cross-entropy loss $\mathcal{L}_{\text{CE}}^l$ is defined as:
\begin{equation}
    \mathcal{L}_{\text{CE}}^l = \frac{1}{|\mathcal{D}_r|}
    \sum_{(x, y) \in \mathcal{D}_r} \text{CE}\big( g_\phi^l(a_\psi^l(h_\theta^l(x))),\, y \big).
    \label{eq:CE_loss}
\end{equation}

\begin{table*}[t]
\centering
\scriptsize
\resizebox{\textwidth}{!}{
\begin{tabular}{l cccc ccc cc c}
\toprule
& \multicolumn{4}{c}{\textbf{ImageNet-1K}} & \multicolumn{3}{c}{\textbf{k-NN}} & & & \\
\cmidrule(lr){2-5} \cmidrule(lr){6-8}
\textbf{Method} & \textbf{FA}$\downarrow$ & \textbf{RA}$\uparrow$ & \textbf{TFA}$\downarrow$ & \textbf{TRA}$\uparrow$ & \textbf{Office-Home} (\textcolor{blue}{$\downarrow$}) & \textbf{CUB} (\textcolor{blue}{$\downarrow$}) & \textbf{DomainNet-126} (\textcolor{blue}{$\downarrow$}) & \textbf{CKA}$\downarrow$ & \textbf{$|$IDI$|$} $\downarrow$ & \textbf{H-Mean}$\uparrow$ \\
\midrule
\textbf{Original}     & 78.98 & 80.01 & 76.10 & 76.47 & 80.28 (\textcolor{blue}{1.95}) & 43.00 (\textcolor{blue}{2.08}) & 72.67 (\textcolor{blue}{10.44}) & 100 & 1.000 & -- \\
\textbf{Retrained}    & 0.00 & 84.47 & 0.00 & 77.62 & 78.33 (\textcolor{blue}{0.00}) & 40.92 (\textcolor{blue}{0.00}) & 83.11 (\textcolor{blue}{0.00}) & 86.19 & 0.000 & -- \\
\midrule
\textbf{PL}                 & 0.61 & 79.46 & 0.42 & 75.59 & \textbf{78.21 (\textcolor{blue}{0.12})} & 44.44 (\textcolor{blue}{3.52}) & 83.73 (\textcolor{blue}{0.62}) & 96.01 & 0.778 & 24.19 \\
\textbf{DUCK}               & 0.04 & 71.21 & 0.02 & 72.34 & 78.67 (\textcolor{blue}{0.34}) & 37.96 (\textcolor{blue}{2.96}) & 81.49 (\textcolor{blue}{1.62}) & 90.15 & 0.538 & 44.65 \\
\textbf{SCAR}               & 5.23 & 79.01 & 4.76 & 77.21 & 80.50 (\textcolor{blue}{2.17}) & 45.25 (\textcolor{blue}{4.33}) & 83.78 (\textcolor{blue}{0.67}) & 96.95 & 0.774 & 20.02 \\
\textbf{SCRUB}              & 1.19 & 67.54 & 1.10 & 65.68 & 76.38 (\textcolor{blue}{1.95}) & 46.48 (\textcolor{blue}{5.56}) & \textbf{82.83 (\textcolor{blue}{0.28})} & 52.60 & 0.702 & 66.31 \\
\textbf{SalUn}              & 23.27 & 39.84 & 21.26 & 35.89 & 46.44 (\textcolor{blue}{31.89}) & 9.97 (\textcolor{blue}{30.95}) & 50.64 (\textcolor{blue}{32.47}) & 9.10 & 0.421 & 59.63 \\
\textbf{RL}                 & 4.31 & 9.56 & 3.76 & 8.98 & 41.74 (\textcolor{blue}{36.59}) & 6.96 (\textcolor{blue}{33.96}) & 46.44 (\textcolor{blue}{36.67}) & \textbf{3.39} & 0.508 & 28.65 \\
\textbf{DELETE}             & 1.58 & \textbf{80.12} & 1.22 & \textbf{77.24} & 79.24 (\textcolor{blue}{0.91}) & 43.94 (\textcolor{blue}{3.02}) & 83.52 (\textcolor{blue}{0.41}) & 97.19 & 0.726 & 19.21 \\
\textbf{COLA}               & \textbf{0.00} & 72.57 & \textbf{0.00} & 73.77 & 78.90 (\textcolor{blue}{0.57}) & \textbf{38.63 (\textcolor{blue}{2.29})} & 81.01 (\textcolor{blue}{2.10}) & 89.28 & 0.867 & 36.54 \\
\textbf{CU}                 & \textbf{0.00} & 75.83 & \textbf{0.00} & 75.49 & 80.62 (\textcolor{blue}{2.29}) & 50.42 (\textcolor{blue}{9.50}) & 83.73 (\textcolor{blue}{0.62}) & 69.52 & 0.403 & 70.68 \\
\midrule
\textbf{EC}               & \textbf{0.00} & 72.63 & \textbf{0.00} & 73.84 & 76.83 (\textcolor{blue}{1.50}) & 44.95 (\textcolor{blue}{4.03}) & 80.88 (\textcolor{blue}{2.23}) & 38.68 & \textbf{0.051} & \textbf{85.75} \\
\bottomrule
\end{tabular}
}
\caption{Comparison of EC against unlearning baselines on ImageNet-1K (ResNet-50, random 100 class forgetting). We report logit-based metrics (Forget Accuracy (FA), Retain Accuracy (RA), Test Forget Accuracy (TFA), Test Retain Accuracy (TRA)), representation-based metrics (CKA, $|$IDI$|$), and k-NN accuracy on three downstream datasets (Office-Home, CUB, DomainNet-126). $|$IDI$|$ denotes the absolute value of IDI. For k-NN results, \textcolor{blue}{blue numbers} indicate the absolute gap relative to the Retrained baseline. H-Mean denotes the harmonic mean of normalized overall metrics. Lower FA, TFA, CKA, and $|$IDI$|$ indicate stronger forgetting, while higher RA and TRA indicate better utility preservation. For k-NN, a smaller absolute gap (w.r.t. the Retrained baseline) indicates better performance. \textbf{Bold} indicates the best performance among the methods.}
\label{tab:tableA}
\end{table*}

\noindent\textbf{Overall Objectives.}
Consequently, the total loss aggregated across all layers is formulated as:
\begin{equation}
    \mathcal{L}_{\text{total}} = \sum_{l=1}^{L} w_l \left( 
    \lambda_{\text{CU}} \mathcal{L}_{\text{CU}}^l + \lambda_{\text{CE}} \mathcal{L}_{\text{CE}}^l 
    \right),
    \label{eq:total_loss}
\end{equation}
where $\lambda_{\text{CU}}$ and $\lambda_{\text{CE}}$ are scaling factors that balance the two loss terms. The layer-wise weights $w_l$ control the contribution of each layer's supervision to the overall objective.
In our experimental settings, we use $L=4$ supervision points for both ResNet-50 and Swin-Tiny and set $(w_1, w_2, w_3, w_4)=(0.2,\,0.4,\,0.8,\,1.0)$. We also set $\lambda_{\text{CU}}=\lambda_{\text{CE}}=1.5$ in all experiments. A hyperparameter sensitivity analysis is included in the Technical Appendix. The overall procedure for EC is summarized in Algorithm~\ref{alg:EC}. 

\section{Experiments}

\subsection{Experimental Setup}
\label{subsec:experimental_setup}


\noindent\textbf{Datasets and Settings.}
We conduct experiments on ImageNet-1K~\cite{deng2009imagenet} and CIFAR-100~\cite{krizhevsky2009cifar} using ResNet-50~\cite{he2016resnet} and Swin-Tiny~\cite{liu2021swin} architectures. For ImageNet-1K, we consider two class-wise forgetting scenarios: \textit{Random-100}, where 100 classes are randomly selected as the forget set, and \textit{Top-100}, where 100 classes most similar to downstream datasets are designated for forgetting following~\cite{kim2025arewe}. For CIFAR-100, we randomly select 10 classes as the forget set and retain the remaining 90 classes. The main text presents results on ResNet-50 under the ImageNet-1K \textit{Random-100} and CIFAR-100 scenarios. Additional experiments on Swin-Tiny and the ImageNet-1K \textit{Top-100} scenario are provided in the Technical Appendix.

\begin{table}[t]
\centering
\scriptsize
\resizebox{\columnwidth}{!}{%
\begin{tabular}{l cccc cc c}
\toprule
& \multicolumn{4}{c}{\textbf{CIFAR-100}} & & & \\
\cmidrule(lr){2-5}
\textbf{Method} & \textbf{FA}$\downarrow$ & \textbf{RA}$\uparrow$ & \textbf{TFA}$\downarrow$ & \textbf{TRA}$\uparrow$ & \textbf{CKA}$\downarrow$ & \textbf{$|$IDI$|$}$\downarrow$ & \textbf{H-Mean}$\uparrow$ \\
\midrule
\textbf{Original}      & 97.04 & 97.47 & 72.90 & 75.26 & 100.00 & 1.000 & -- \\
\textbf{Retrained}     & 0.00  & 97.51 & 0.00  & 75.88 & 72.59  & 0.000 & -- \\
\midrule
\textbf{PL}            & 4.60  & 97.49 & 4.00  & 75.56 & 82.17  & 0.853 & 35.64 \\
\textbf{DUCK}          & 4.56  & 98.76 & 4.70  & 73.87 & 83.39  & 0.763 & 40.81 \\
\textbf{SCAR}          & 0.29  & 97.60 & 0.70  & \textbf{76.67} & 88.72  & 0.685 & 36.63 \\
\textbf{SCRUB}         & 0.02  & 97.18 & \textbf{0.00}  & 73.96 & 75.78  & 0.577 & 55.17 \\
\textbf{SalUn}         & 44.93 & 97.77 & 24.30 & 75.54 & 59.33  & 0.873 & 37.97 \\
\textbf{RL}            & 31.47 & 97.99 & 19.00 & 75.92 & \textbf{44.66} & 0.927 & 29.24 \\
\textbf{DELETE}   & 1.49  & 97.54 & 0.50  & 75.68 & 83.26  & 0.822 & 37.56 \\
\textbf{COLA}     & \textbf{0.00}  & \textbf{99.57} & \textbf{0.00}  & 75.13 & 79.11  & 0.744 & 46.06 \\
\textbf{CU}            & 0.09  & 97.20 & \textbf{0.00}  & 76.22 & 66.93  & 0.539 & 62.94 \\
\midrule
\textbf{EC} & \textbf{0.00}  & 95.63 & \textbf{0.00}  & 74.80 & 61.98 & \textbf{0.291} & \textbf{71.23} \\
\bottomrule
\end{tabular}
}
\caption{Comparison of unlearning methods on \textbf{CIFAR-100} (ResNet-50, random 10 class forgetting). We exclude k-NN transfer results on downstream datasets for CIFAR-100 since the transfer accuracy is near chance (lower than \(\sim 5\%\)) even for the original model, making it uninformative for comparing methods. Metric definitions and comparison protocol follow Table~\ref{tab:tableA}; \textbf{bold} denotes the best among methods.}
\label{tab:table_cifar100}
\end{table}

\noindent\textbf{Baselines.}
We compare our proposed method against the Original model and the Retrained model (the gold standard), as well as several state-of-the-art unlearning baselines: Pseudo Labeling (PL)~\cite{chen2024unsc}, DUCK~\cite{cotogni2023duck}, SCAR~\cite{bonato2024retain}, SCRUB~\cite{kurmanji2023towards}, SalUn~\cite{fan2024salun}, Random Labeling (RL)~\cite{golatkar2020eternal}, DELETE~\cite{zhou2025decoupled}, COLA~\cite{jeon2024idi} and Contrastive Unlearning (CU)~\cite{zhang2024contrastive}.

\begin{figure*}[t]
    \centering
    \includegraphics[width=1.0\linewidth]{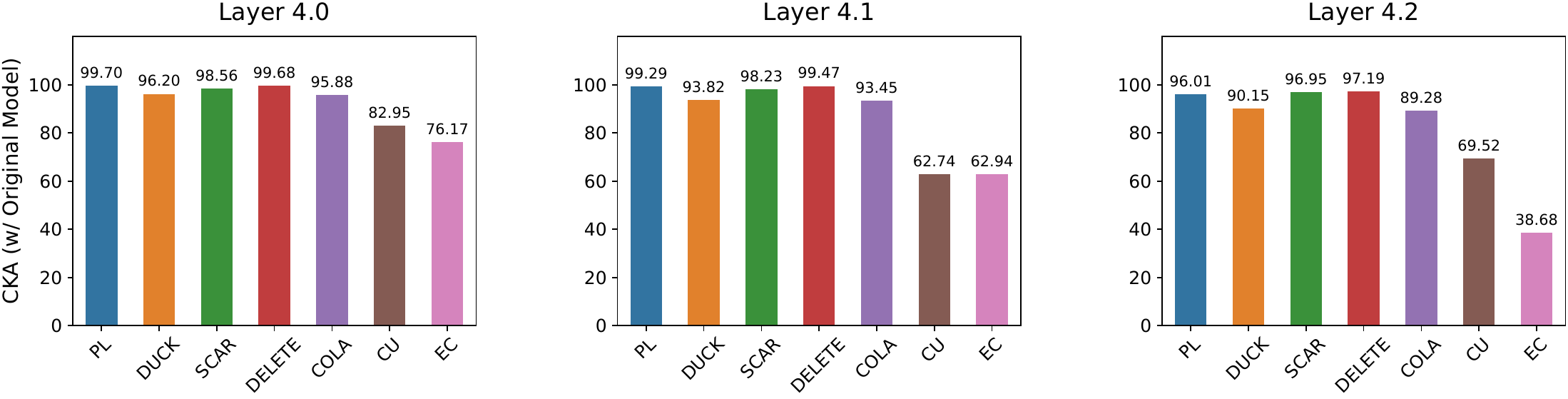}
    \caption{Layer-wise representational similarity to the original model measured by CKA on the test forget set. We compare features from ResNet-50 Layers 4 (=4.2) and two Layer-4 bottleneck blocks (Layer 4.0, 4.1); lower CKA indicates larger deviation from the original model.}
    \label{fig:cka barplots figure}
\end{figure*}

\noindent\textbf{Evaluation Metrics.}
We employ logit-based metrics to assess the effectiveness of unlearning. Specifically, we report Forget Accuracy (FA) on $\mathcal{D}_f$ and Test Forget Accuracy (TFA) on $\mathcal{D}_f^{\text{te}}$ to measure the degree of forgetting. To evaluate the preservation of model utility, we report Retain Accuracy (RA) on $\mathcal{D}_r$ and Test Retain Accuracy (TRA) on $\mathcal{D}_r^{\text{te}}$.

For representation-based evaluation, we adopt Centered Kernel Alignment (CKA)~\cite{kornblith2019similarity}. We measure the similarity between the original model $f_o$ and the unlearned model $f_u$ using features extracted from the test forget set $\mathcal{D}_f^{\text{te}}$. Unless otherwise specified, CKA values reported in tables are computed using the penultimate-layer features (i.e., the final layer output before the classifier). In our experiments with ResNet-50, this corresponds to Layer 4 (denoted as Layer 4.2), where high-level, class-discriminative features are encoded. To examine representational changes at finer granularity, we additionally compute CKA at two intermediate bottleneck blocks within Layer 4 (Layer 4.0 and Layer 4.1) for layer-wise analysis in Figure~\ref{fig:cka barplots figure}.

Furthermore, we adopt the Information Difference Index (IDI)~\cite{jeon2024idi} for representation-based evaluation. Following the original implementation, we compute IDI using the last three bottleneck blocks of Layer 4 in ResNet-50. IDI is normalized such that a value of 0 indicates complete unlearning equivalent to the retrained model, while 1 indicates no unlearning (equivalent to the original model). Negative values indicate over-unlearning, and values exceeding 1 suggest the model retains more forget-set information than the original.

In addition to CKA and IDI, we evaluate representational quality through downstream task performance. Following \citeauthor{kim2025arewe}~\shortcite{kim2025arewe}, we employ k-nearest neighbor (k-NN) classification~\cite{cover1967} on three downstream datasets: Office-Home~\cite{venkateswara2017}, CUB-200-2011~\cite{catherineundefined}, and DomainNet-126~\cite{peng2018}. We freeze the backbone of each unlearned model and train a k-NN classifier on the extracted features. This assesses whether representations retain transferable structure despite unlearning.

To provide a unified measure that balances forgetting effectiveness and utility preservation, we compute the harmonic mean across all evaluation metrics. Since each metric has a different orientation and scale, we first normalize them to a common scale where higher values indicate better performance. For metrics where lower is better (FA, TFA, CKA), we subtract the value from 100. For metrics where higher is better (RA, TRA), we use the raw values directly. For k-NN downstream performance, we subtract the absolute gap from the Retrained baseline from 100. For $|\text{IDI}|$, which ranges from 0 to 1, we subtract the value from 1 and scale by 100. The harmonic mean is then computed over all nine normalized scores.

\subsection{Experimental Results}

\textbf{Large-scale Multi-Class Unlearning.}
Tables~\ref{tab:tableA} and~\ref{tab:table_cifar100} present comparisons on ImageNet-1K (random 100 class forgetting) and CIFAR-100 (random 10 class forgetting), respectively. We exclude downstream k-NN results for CIFAR-100 as transfer accuracy is near chance level ($\sim$5\%) even for the Original model.

The results show that EC achieves strong representation-level forgetting across both benchmarks. While maintaining reasonable Test Retain Accuracy (TRA), EC attains the lowest CKA among utility-preserving baselines (excluding SalUn and RL, which exhibit severe utility degradation), the smallest $|\text{IDI}|$, and the best H-Mean among all methods. These results confirm that EC effectively induces changes in intermediate representations while preserving retain set performance.

\noindent\textbf{Intermediate Layer CKA.}
To examine the extent of unlearning at intermediate layers, we measure the representational similarity between the original model and the unlearned model using CKA. Specifically, we compute CKA using the output features from Layer 4 and its intermediate bottleneck blocks (\ie, Layer 4.0 and Layer 4.1). Notably, the features are extracted directly from the backbone, independent of the EC modules.

The results are visualized in Figure~\ref{fig:cka barplots figure}. Most baseline methods---including PL, DUCK, SCAR, DELETE, and COLA---show consistently high similarity with the original model across all layers. While SCRUB, SalUn, and RL achieve low similarity scores, they incur substantial degradation in retain set performance (Table~\ref{tab:tableA}). Therefore, we exclude these methods from the subsequent baseline comparison and focus on approaches that maintain reasonable retain utility. Among the remaining baselines, CU achieves the lowest similarity while maintaining reasonable retain accuracy. Our proposed method, EC, attains even lower similarity scores than CU across all layers. In particular, the reduction in similarity is most pronounced in the later layers (from layer 4.0 onward), where high-level, class-discriminative features are encoded.

\begin{figure}[t]
    \centering
    \begin{subfigure}[b]{0.48\linewidth}
        \centering
        \includegraphics[width=\linewidth]{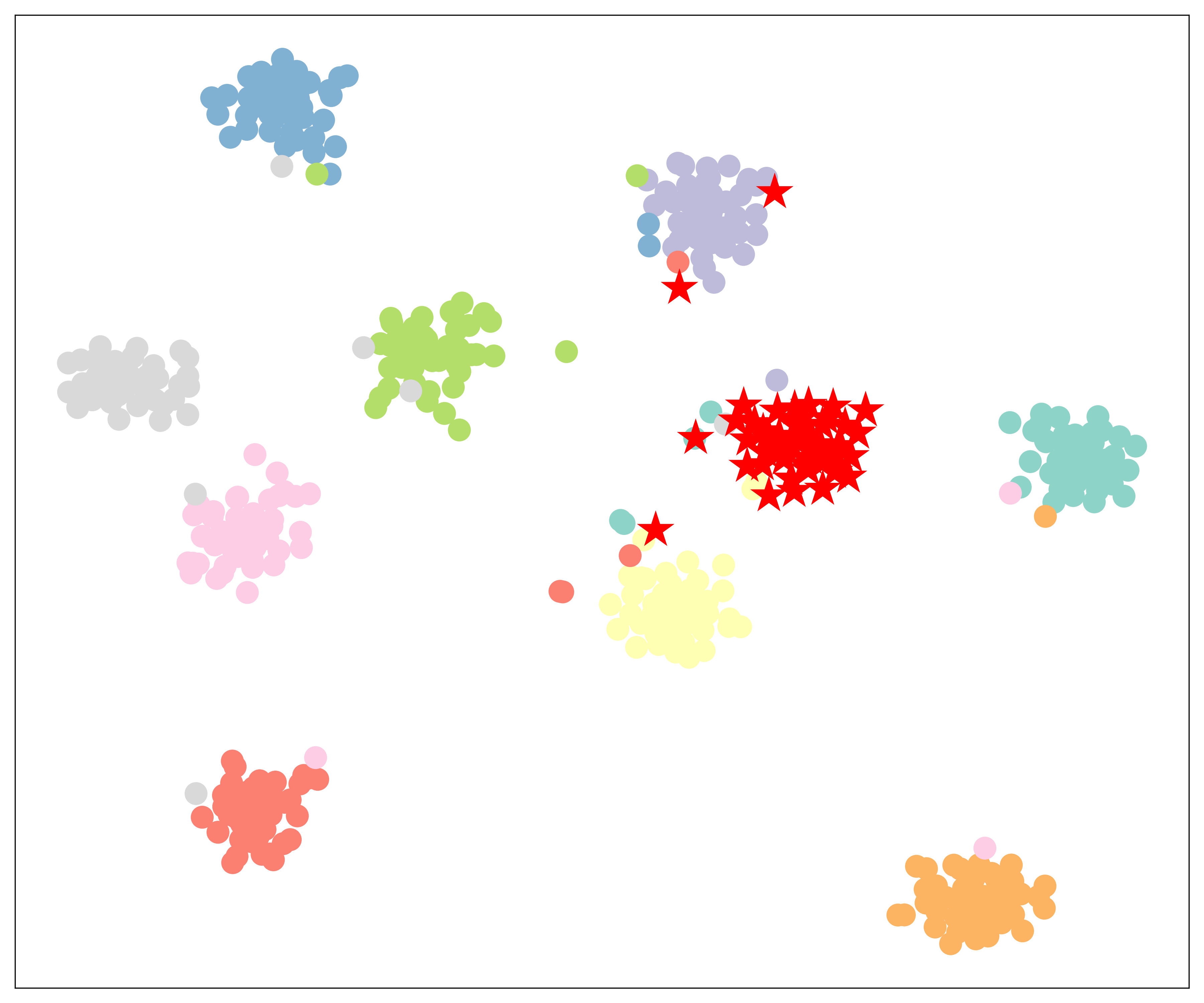}
        \caption{Original}
        \label{fig:tsne_original}
    \end{subfigure}
    \begin{subfigure}[b]{0.48\linewidth}
        \centering
        \includegraphics[width=\linewidth]{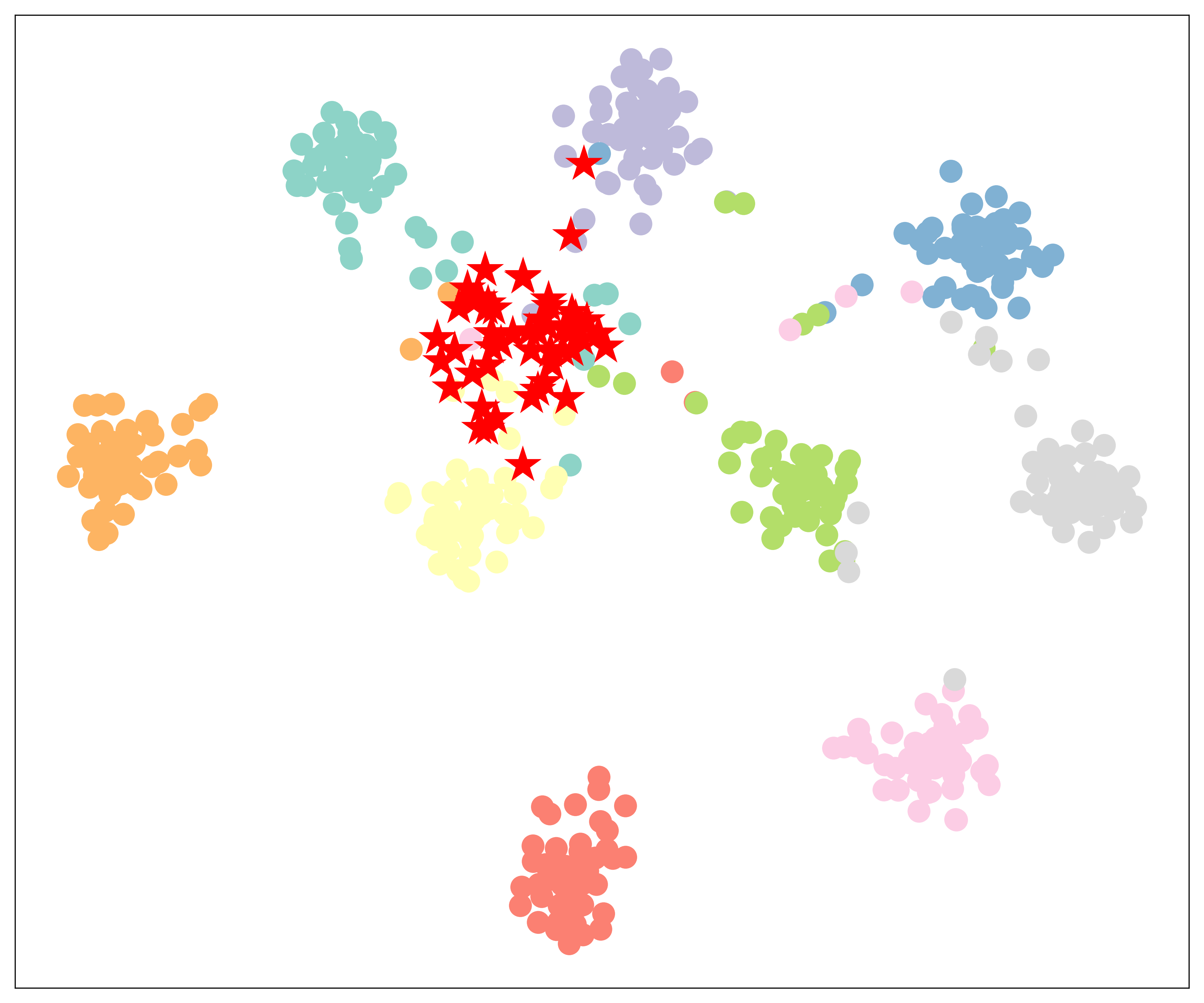}
        \caption{Retrained}
        \label{fig:tsne_retrained}
    \end{subfigure}

    \begin{subfigure}[b]{0.48\linewidth}
        \centering
        \includegraphics[width=\linewidth]{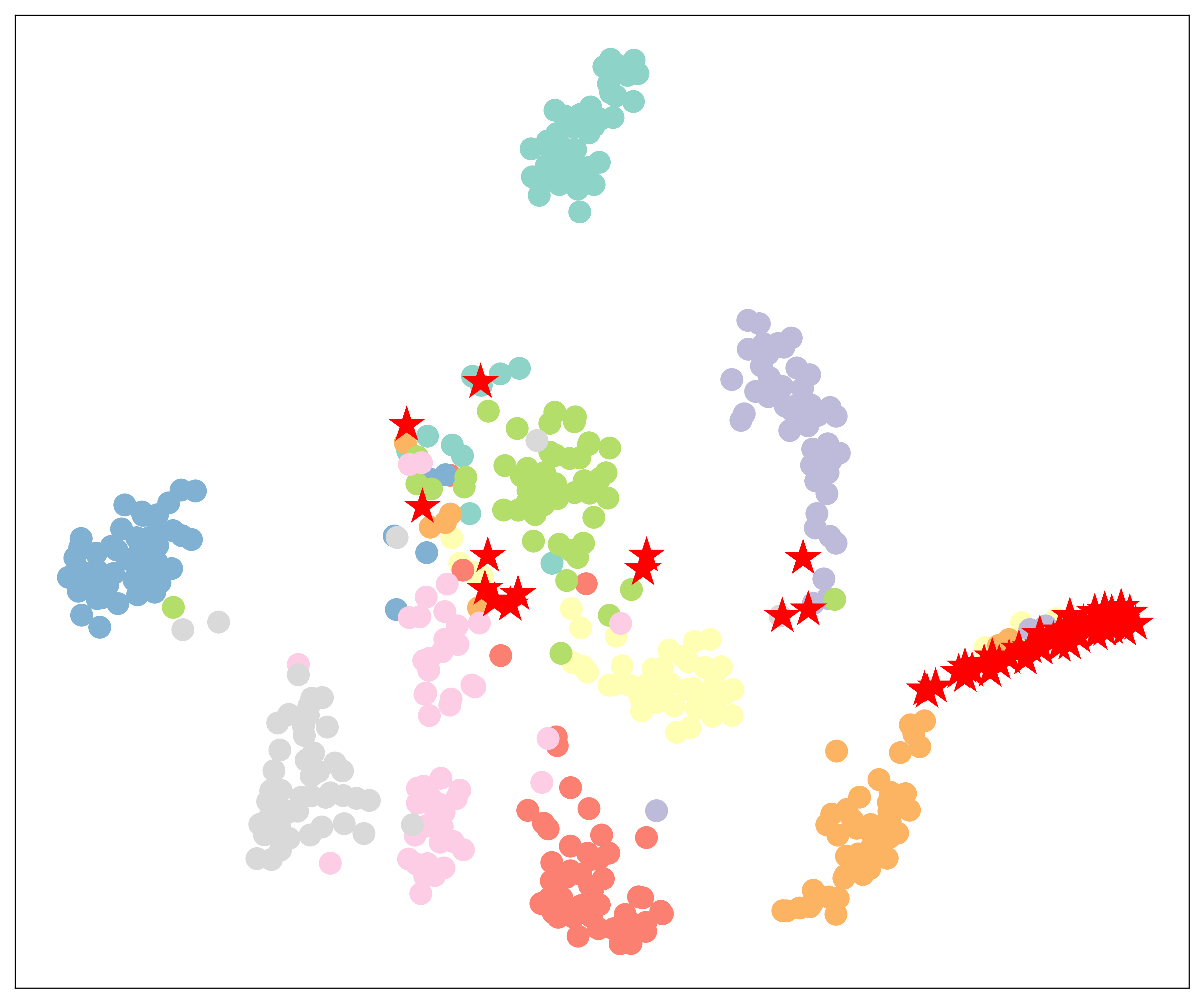}
        \caption{CU}
        \label{fig:tsne_cu}
    \end{subfigure}
    \begin{subfigure}[b]{0.48\linewidth}
        \centering
        \includegraphics[width=\linewidth]{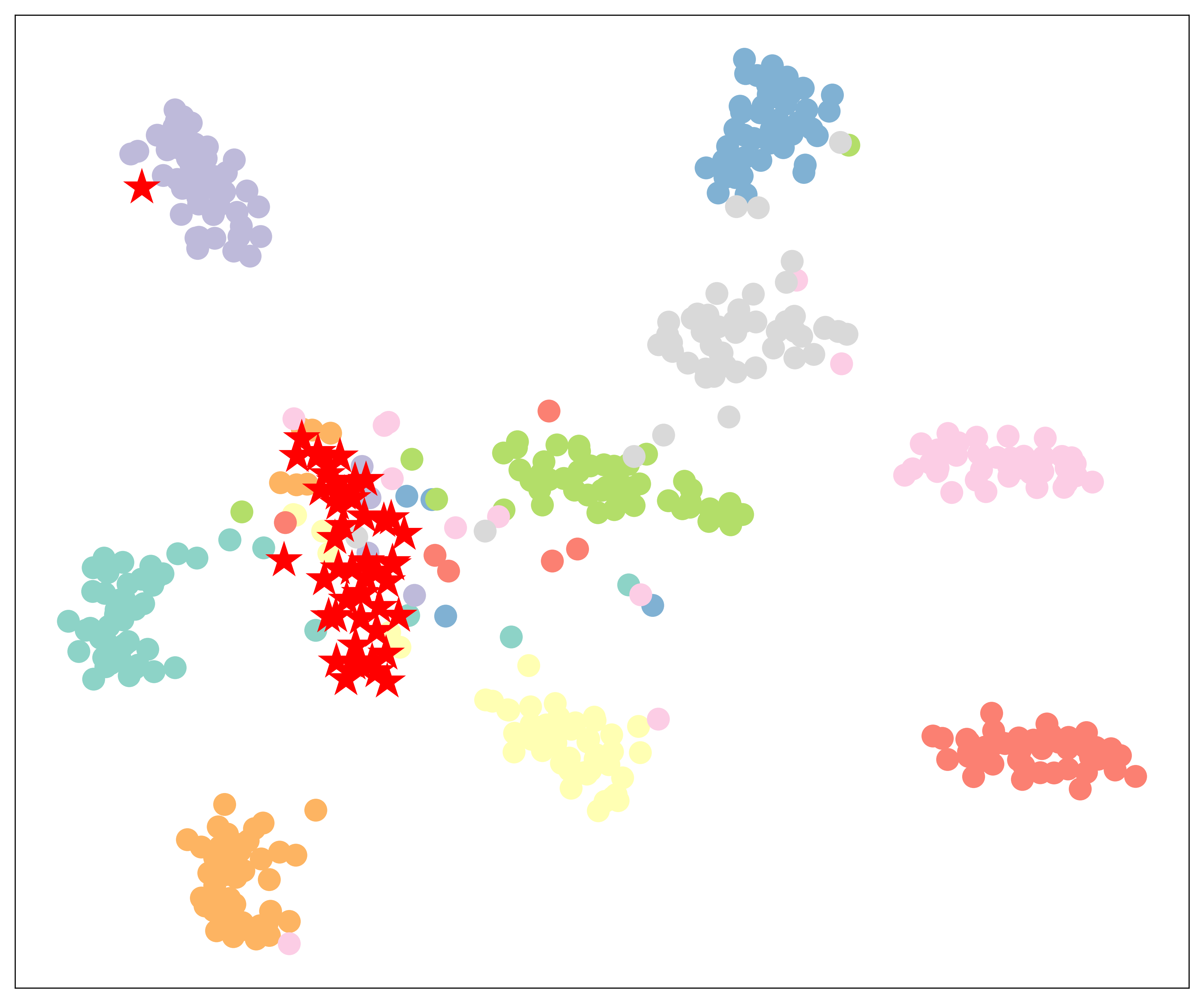}
        \caption{EC}
        \label{fig:tsne_EC}
    \end{subfigure}

    \caption{t-SNE visualization of the pooled feature representation before the final classifier.
    Red stars are the forget class, circles are 9 retain classes with the highest similarity to the forget class.}
    \label{fig:tsne}
\end{figure}

\noindent\textbf{Representation Visualization.}
To qualitatively inspect how unlearning reshapes the feature space, we visualize feature representations using t-SNE~\cite{vandermaaten2008tsne}. Figure~\ref{fig:tsne} compares embeddings produced by the Original model, the Retrained model, CU, and the EC.

In the Original model, forget classes form compact, well-separated clusters, reflecting strong class-discriminative structure. In the Retrained model, the forget class features become more dispersed, and the boundaries between forget classes and nearby retain classes are noticeably weakened. Similarly, EC exhibits a distribution qualitatively close to the Retrained baseline: forget class embeddings spread out and the separation from surrounding retain class regions largely collapses, indicating that the class-discriminative structure is degraded beyond logit-level effects.

\begin{table}[t]
\centering
\scriptsize
\begin{tabular}{lcccccc}
\toprule
\textbf{Method} & \textbf{FA}$\downarrow$ & \textbf{RA}$\uparrow$ & \textbf{TFA}$\downarrow$ & \textbf{TRA}$\uparrow$ & \textbf{CKA}$\downarrow$ & \textbf{$|$IDI$|$}$\downarrow$\\
\midrule
\textbf{CU}        & \textbf{0.00} & \textbf{75.83} & \textbf{0.00} & \textbf{75.49} & 69.52 & 0.403 \\
\textbf{EC}      
                   & \textbf{0.00} & 72.63 & \textbf{0.00} & 73.84 & \textbf{38.68} & \textbf{0.051} \\
\midrule
\textbf{DUCK}      & 0.04 & 71.21 & 0.02 & 72.34 & 90.15 & 0.538 \\
\textbf{DUCK (+EC)}
                   & \textbf{0.00} & \textbf{72.02} & \textbf{0.00} & \textbf{73.00} & \textbf{87.67} & \textbf{0.511} \\
\midrule
\textbf{COLA}      & \textbf{0.00} & \textbf{72.57} & \textbf{0.00} & \textbf{73.77} & 89.28 & 0.867 \\
\textbf{COLA (+EC)}
                   & \textbf{0.00} & 72.43 & \textbf{0.00} & 73.75 & \textbf{89.15} & \textbf{0.846} \\
\bottomrule
\end{tabular}
\caption{Comparison of baselines with and without EC on ImageNet-1K (ResNet-50, random 100 class forgetting). (+EC) denotes the application of EC's multi-layer supervision to each baseline. The better performance in each comparison is shown in \textbf{bold}.}
\label{tab:tableB}
\end{table}

\noindent\textbf{EC Combined with Other Unlearning Baselines.}
To verify that EC can serve as a model-agnostic plug-in module, we apply EC to other representation-based unlearning baselines and evaluate their EC-augmented variants. Table~\ref{tab:tableB} presents a comparison between the CU and EC with DUCK, COLA and their EC-augmented counterparts (DUCK (+EC), COLA (+EC)).
Overall, integrating EC tends to improve representation-based forgetting metrics (CKA and IDI), while utility metrics (RA/TRA) are maintained or slightly improved depending on the baseline. Notably, DUCK (+EC) improves both RA and TRA over DUCK, indicating that EC can enhance representation-level forgetting while maintaining—or even improving—retain set utility; however, its utility impact can be baseline-dependent (e.g., EC shows a drop in RA/TRA relative to CU).

\subsection{Ablation Study}

We conduct ablation studies to analyze the contribution of each component in EC. Table~\ref{tab:ablation} presents the ResNet-50 results on ImageNet-1K under the random 100 class forgetting setting.

Removing layer-wise cross-entropy loss (w/o layer-wise CE) leads to higher CKA and IDI, indicating that intermediate CE supervision contributes to both utility preservation and effective representation-level forgetting.

Removing EC modules (w/o EC modules) results in unstable optimization with negative IDI (raw IDI= -0.233), indicating over-forgetting. This suggests that EC modules serve as stabilizing adapters that prevent overly aggressive representational drift.

Extending supervision to additional bottleneck blocks within final backbone blocks (in ResNet-50, the first two bottleneck blocks of Layer 4, i.e., layer4.0 and layer4.1) yields slightly better retain set utility but weaker forgetting. This trade-off suggests that overlapping supervision signals within the same layer block may constrain the model's capacity to reorganize its representations.

\begin{table}[t]
\centering
\scriptsize
\resizebox{0.48\textwidth}{!}{
\begin{tabular}{lcccccc}
\toprule
\textbf{Method} & \textbf{FA}$\downarrow$ & \textbf{RA}$\uparrow$ & \textbf{TFA}$\downarrow$ & \textbf{TRA}$\uparrow$ & \textbf{CKA}$\downarrow$ & \textbf{$|$IDI$|$}$\downarrow$ \\
\midrule
\textbf{CU}                     & 0.00 & 75.83 & 0.00 & 75.49 & 69.52 & 0.403 \\
\midrule
\textbf{EC}                  & \textbf{0.00} & 72.63 & \textbf{0.00} & 73.84 & \textbf{38.68} & \textbf{0.051} \\
\quad w/o layer-wise ce        & \textbf{0.00} & \textbf{74.66} & \textbf{0.00} & 74.40 & 59.28 & 0.205 \\
\quad w/o EC modules           & \textbf{0.00} & 71.37 & \textbf{0.00} & 73.48 & 63.07 & 0.233 \\
\quad +final backbone blocks      & \textbf{0.00} & 73.48 & \textbf{0.00} & \textbf{74.54} & 56.03 & 0.100 \\
\bottomrule
\end{tabular}
}
\caption{Ablation study on ImageNet-1K (ResNet-50, random 100 class forgetting). ``+final backbone blocks'' denotes additional supervision at the last backbone blocks
(in ResNet-50, the first two bottleneck blocks of Layer 4, i.e., Layer 4.0 and Layer 4.1). $|$IDI$|$ denotes the absolute value of IDI. \textbf{Bold} denotes the best performance across ablation variants.}
\label{tab:ablation}
\end{table}

\section{Conclusion}
We addressed the problem of superficial forgetting in approximate machine unlearning, where models exhibit low forget set accuracy yet retain substantial residual information in their internal representations. To overcome this limitation, we introduced Erase at the Core (EC), a framework that enforces forgetting across the depth of the network through multi-layer contrastive unlearning and deep supervision.

Across multiple benchmarks---including ImageNet-1K (both random and top-100 class forgetting scenarios) and CIFAR-100---and architectures (ResNet-50 and Swin-Tiny), EC achieves strong representation-level forgetting, attaining the highest harmonic mean score. Furthermore, EC serves as a model-agnostic plug-in module that consistently improves representation-based forgetting when applied to baselines such as DUCK and COLA, while maintaining retain set utility.

Our evaluation provides strong empirical evidence of representation-level forgetting; formal erasure guarantees remain an open direction for future work. Multi-layer supervision also introduces additional computational overhead. Future work includes extending EC to more diverse architectures and integrating it with additional unlearning methods.

\newpage
\bibliographystyle{named}
\bibliography{references}

\clearpage
\appendix
\section*{Technical Appendix}
\addcontentsline{toc}{section}{Technical Appendix} 










This appendix provides supplementary experiments and implementation details that complement the main results (ResNet-50 on ImageNet-1K under the Random-100 setting). Section~\ref{sec:additional_results} presents additional experimental results, including qualitative visualization, alternative forgetting scenarios, datasets, and architectures. Section~\ref{sec:experimental_details} describes implementation details for reproducibility.

\section{Additional Results}
\label{sec:additional_results}


\subsection{Qualitative Analysis via k-NN Retrieval Visualization}
\label{sec:knn_visualization}
\begin{figure*}[t]
    \centering
    \includegraphics[width=1.0  \linewidth]{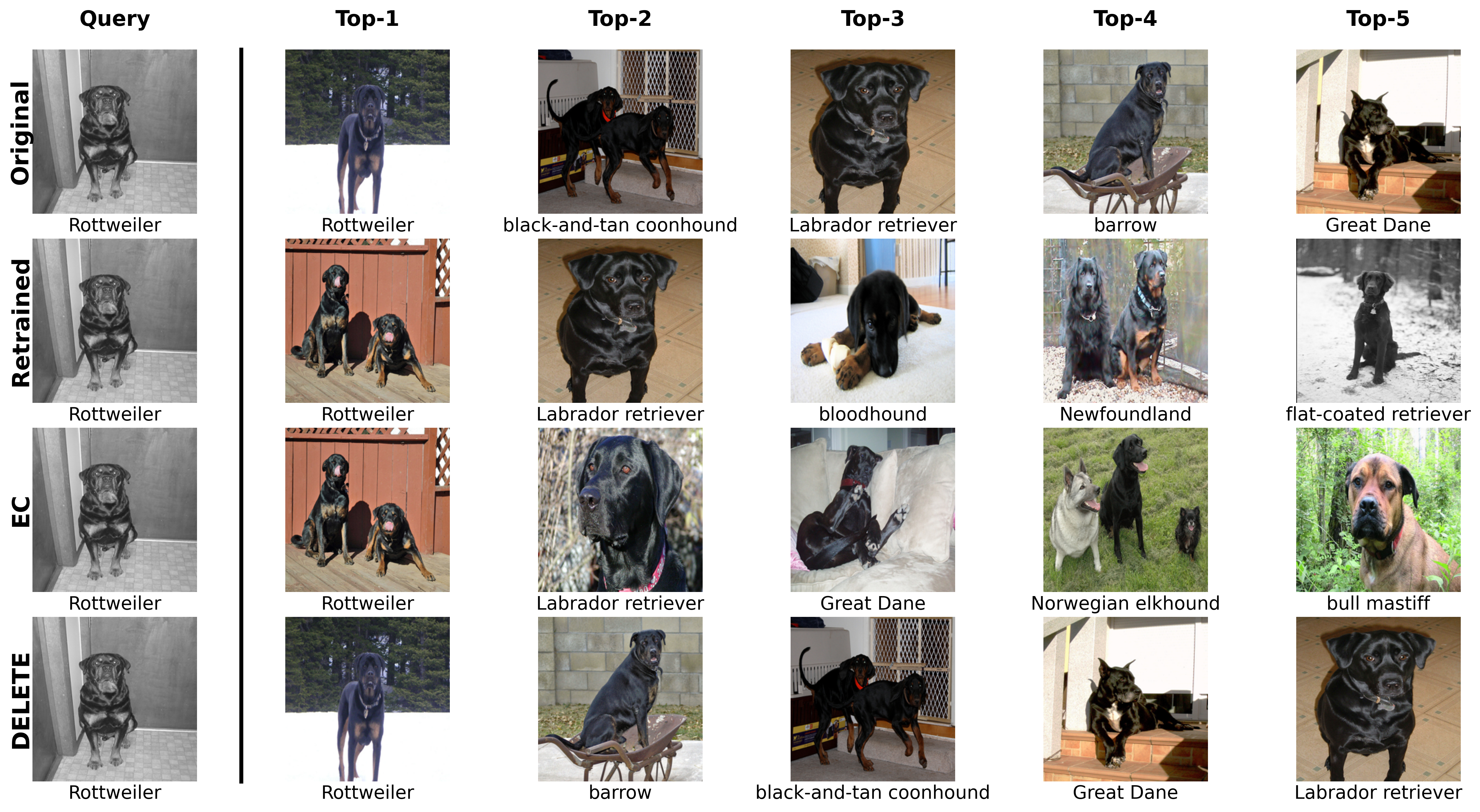}
    \caption{Comparison of k-NN retrieval results for a query image from the forget set under the ImageNet-1K, ResNet-50, random 100 class forgetting setup. All four models retrieve the same class (Rottweiler) at top-1, but the Retrained and EC models return the same retrieved image, while the Original model and DELETE return a different one. DELETE retrieves the same images as the Original model across top-5, differing only in order, indicating minimal representation-level change. From top-2 onward, the EC model's retrievals align more closely with those of the Retrained model than with the Original model or DELETE.}
    \label{fig:knn_retrieval}
\end{figure*}

Figure~\ref{fig:knn_retrieval} visualizes the top-$k$ nearest-neighbor retrievals for a query image from the forget set under the ImageNet-1K / ResNet-50 / Random-100 setting. We compare the retrieved results from the Original model, the Retrained baseline, DELETE~\cite{zhou2025decoupled}, and EC. While all four models retrieve the same class with query (Rottweiler) as the top-1 result, the Retrained model and EC return the same image, whereas the Original model and DELETE retrieve a different one. Notably, DELETE produces the identical top-1 retrieval image as the Original model, indicating that its representations remain closely aligned with the pre-unlearning state. Furthermore, across the top-5 retrievals, DELETE returns the same set of classes and images as the Original model, differing only in the order. This suggests that DELETE primarily modifies output-level predictions without inducing substantial changes in the underlying feature representations. At top-2, both the Retrained model and EC retrieve the same class (Labrador retriever), while the Original model retrieve a different class (black-and-tan coonhound). These observations indicate that EC produces retrieval results that are qualitatively closer to the Retrained model than to the Original model, demonstrating that EC induces meaningful representation-level changes away from the Original model, whereas DELETE does not.

\subsection{Top-100 Class Forgetting on ImageNet-1K}
\label{sec:top100}
Table~\ref{tab:table_top100} reports results on the Top-100 class forgetting scenario, where the forget classes are selected based on their semantic similarity to a downstream dataset (CUB-200-2011). This setting is more challenging than Random-100 because the forget classes overlap semantically with the downstream task, requiring substantial divergence of feature representations to prevent information leakage. EC achieves the highest H-Mean score (69.98) while maintaining the lowest CKA (77.89) among methods that preserve reasonable retain utility. These results demonstrate that EC remains effective under this more challenging scenario.

\begin{table*}[t]
\centering
\scriptsize
\resizebox{\textwidth}{!}{

\begin{tabular}{l cccc ccc cc c}
\toprule
& \multicolumn{4}{c}{\textbf{ImageNet-1K}} & \multicolumn{3}{c}{\textbf{k-NN}} & & & \\
\cmidrule(lr){2-5} \cmidrule(lr){6-8}
\textbf{Method} & \textbf{FA}$\downarrow$ & \textbf{RA}$\uparrow$ & \textbf{TFA}$\downarrow$ & \textbf{TRA}$\uparrow$ & \textbf{Office-Home} (\textcolor{blue}{$\downarrow$}) & \textbf{CUB} (\textcolor{blue}{$\downarrow$}) & \textbf{DomainNet} (\textcolor{blue}{$\downarrow$}) & \textbf{CKA}$\downarrow$ & \textbf{$|$IDI$|$} $\downarrow$ & \textbf{H-Mean}$\uparrow$ \\
\midrule
\textbf{Original}     & 82.85 & 79.63 & 81.04 & 75.92 & 80.28 (\textcolor{blue}{0.22}) & 43.00 (\textcolor{blue}{20.18}) & 72.67 (\textcolor{blue}{9.08}) & 100.00 & 1.000 & -- \\
\textbf{Retrained}    & 0.00  & 74.89 & 0.00  & 74.27 & 80.50 (\textcolor{blue}{0.00}) & 22.82 (\textcolor{blue}{0.00}) & 81.75 (\textcolor{blue}{0.00}) & 78.65 & 0.000 & -- \\
\midrule
\textbf{PL}           & 0.37 & 78.61 & 0.42 & 75.58 & 79.24 (\textcolor{blue}{1.26}) & 37.23 (\textcolor{blue}{14.41}) & 83.82 (\textcolor{blue}{2.07}) & 93.96 & 0.866 & 28.28 \\
\textbf{DUCK}         & 0.10 & 70.12 & 0.08 & 70.89 & 78.44 (\textcolor{blue}{2.06}) & 29.09 (\textcolor{blue}{6.27}) & 80.83 (\textcolor{blue}{0.92}) & 86.42 & 0.207 & 54.35 \\
\textbf{SCAR}         & 26.42 & 77.95 & 29.92 & \textbf{75.68} & 79.47 (\textcolor{blue}{1.03}) & 40.63 (\textcolor{blue}{17.81}) & 83.69 (\textcolor{blue}{1.94}) & 96.88 & 0.720 & 20.34 \\
\textbf{SCRUB}        & 9.66 & 62.48 & 10.78 & 63.76 & 72.48 (\textcolor{blue}{8.02}) & \textbf{23.54 (\textcolor{blue}{0.72})} & 78.08 (\textcolor{blue}{3.67}) & 64.48 & \textbf{0.031} & 64.12 \\
\textbf{SalUn}        & 17.24 & 49.97 & 12.46 & 44.60 & 62.39 (\textcolor{blue}{18.11}) & 8.10 (\textcolor{blue}{14.72}) & 59.95 (\textcolor{blue}{21.80}) & 38.01 & 1.590 & 41.13 \\
\textbf{RL}           & 10.54 & 39.51 & 7.90 & 34.37 & 49.31 (\textcolor{blue}{31.19}) & 5.68 (\textcolor{blue}{17.14}) & 45.91 (\textcolor{blue}{35.84}) & \textbf{10.47} & 1.659 & 39.17 \\
\textbf{DELETE}       & 9.68 & \textbf{78.73} & 6.70 & 75.45 & 79.47 (\textcolor{blue}{1.03}) & 40.12 (\textcolor{blue}{17.30}) & 83.63 (\textcolor{blue}{1.88}) & 94.18 & 0.842 & 28.57 \\
\textbf{COLA}         & \textbf{0.00} & 71.22 & \textbf{0.00} & 72.38 & 79.93 (\textcolor{blue}{0.57}) & 28.88 (\textcolor{blue}{6.06}) & \textbf{81.52 (\textcolor{blue}{0.23})} & 85.58 & 0.802 & 45.34 \\
\textbf{CU}           & 0.01 & 73.33 & \textbf{0.00} & 73.49 & \textbf{80.39 (\textcolor{blue}{0.11})} & 46.73 (\textcolor{blue}{23.91}) & 82.86 (\textcolor{blue}{1.11}) & 84.69 & 0.311 & 56.13 \\
\midrule
\textbf{EC}           & \textbf{0.00} & 71.50 & \textbf{0.00} & 72.14 & 79.58 (\textcolor{blue}{0.92}) & 45.46 (\textcolor{blue}{22.64}) & 82.07 (\textcolor{blue}{0.32}) & 77.89 & 0.203 & \textbf{69.98} \\
\bottomrule
\end{tabular}
}
\caption{Comparison of unlearning methods on ImageNet-1K (ResNet-50, \textbf{Top 100} class forgetting). Here, the \textbf{Top 100} classes are selected from ImageNet-1K as the 100 classes most similar to the downstream dataset CUB-200-2011. $|$IDI$|$ denotes the absolute value of IDI. For k-NN results, \textcolor{blue}{blue numbers} indicate the absolute gap relative to the Retrained baseline. H-Mean denotes the harmonic mean of normalized overall metrics. Lower FA, TFA, CKA, and $|$IDI$|$ indicate stronger forgetting, while higher RA and TRA indicate better utility preservation. For k-NN, a smaller absolute gap (w.r.t. the Retrained baseline) indicates better performance. \textbf{Bold} indicates the best performance among unlearning methods.}
\label{tab:table_top100}
\end{table*}

\subsection{Additional Backbone: Swin-Tiny on ImageNet-1K}
\label{sec:swin}

Table~\ref{tab:swin_tiny_random100} extends the evaluation to the Swin-Tiny architecture under ImageNet-1K Random-100 forgetting. This experiment verifies that the empirical behavior observed in the main text is not tied to a specific CNN backbone. For Swin-Tiny, we use a variant of EC without layer-wise cross-entropy loss, applying the CE loss only at the final classifier (corresponding to the ``w/o layer-wise ce'' configuration in the ablation study of the main paper). EC achieves the highest H-Mean (62.51) and the lowest $|\text{IDI}|$ (0.477) among methods that maintain reasonable retain utility, supporting the architecture-agnostic nature of attaching EC modules at intermediate stages.

\begin{table*}[t]
\centering
\scriptsize
\resizebox{\textwidth}{!}{
\begin{tabular}{l cccc ccc cc c}
\toprule
& \multicolumn{4}{c}{\textbf{ImageNet-1K}} & \multicolumn{3}{c}{\textbf{k-NN}} & & & \\
\cmidrule(lr){2-5} \cmidrule(lr){6-8}
\textbf{Method} & \textbf{FA}$\downarrow$ & \textbf{RA}$\uparrow$ & \textbf{TFA}$\downarrow$ & \textbf{TRA}$\uparrow$ &
\textbf{Office-Home} (\textcolor{blue}{$\downarrow$}) & \textbf{CUB} (\textcolor{blue}{$\downarrow$}) & \textbf{DomainNet} (\textcolor{blue}{$\downarrow$}) &
\textbf{CKA}$\downarrow$ & \textbf{$|$IDI$|$}$\downarrow$ & \textbf{H-Mean}$\uparrow$ \\
\midrule
\textbf{Original}  & 77.89 & 76.65 & 76.48 & 78.56 & 88.53 (\textcolor{blue}{0.69}) & 74.39 (\textcolor{blue}{0.00}) & 88.58 (\textcolor{blue}{0.21}) & 100.00 & 1.000 & -- \\
\textbf{Retrained} & 0.00  & 79.27 & 0.00  & 79.85 & 87.84 (\textcolor{blue}{0.00}) & 74.39 (\textcolor{blue}{0.00}) & 88.79 (\textcolor{blue}{0.00}) & 89.29  & 0.000 & -- \\
\midrule
\textbf{PL}        & 0.76 & 75.53 & 0.88 & 76.76 & \textbf{86.12 (\textcolor{blue}{1.72})} & 70.82 (\textcolor{blue}{3.57}) & 87.21 (\textcolor{blue}{1.58}) & 89.19 & 0.522 & 47.23 \\
\textbf{DUCK}      & \textbf{0.00} & 64.72 & \textbf{0.00} & 67.87 & 78.10 (\textcolor{blue}{9.74}) & 53.01 (\textcolor{blue}{21.38}) & 81.94 (\textcolor{blue}{6.85}) & 74.90 & 2.695 & 40.08 \\
\textbf{SCAR}      & \textbf{0.00} & 0.11  & \textbf{0.00} & 0.11  & 1.95  (\textcolor{blue}{85.89}) & 0.34  (\textcolor{blue}{74.05}) & 0.80  (\textcolor{blue}{87.99}) & \textbf{2.03}  & 6.132 & 0.49 \\
\textbf{SCRUB}     & 5.46 & 68.32 & 5.76 & 71.12 & 83.72 (\textcolor{blue}{4.12}) & 64.46 (\textcolor{blue}{9.93})  & 86.35 (\textcolor{blue}{2.44}) & 74.48 & 1.460 & 40.75 \\
\textbf{SalUn}     & 6.69 & 6.38  & 7.76 & 6.78  & 67.43 (\textcolor{blue}{20.41})& 55.22 (\textcolor{blue}{19.17}) & 79.04 (\textcolor{blue}{9.75}) & 16.18 & 3.871 & 19.00 \\
\textbf{RL}        & 6.10 & 5.59  & 6.90 & 5.96  & 67.55 (\textcolor{blue}{20.29})& 55.00 (\textcolor{blue}{19.39}) & 78.57 (\textcolor{blue}{10.22})& 13.59 & 3.902 & 17.45 \\
\textbf{DELETE}    & 2.75 & \textbf{75.94} & 2.62 & \textbf{77.15} & 85.89 (\textcolor{blue}{1.95}) & \textbf{72.69 (\textcolor{blue}{1.70})}  & \textbf{87.84 (\textcolor{blue}{0.95})} & 95.91 & 0.718 & 25.20 \\
\textbf{COLA}      & \textbf{0.00} & 67.02 & \textbf{0.00} & 70.78 & 80.28 (\textcolor{blue}{7.56}) & 59.80 (\textcolor{blue}{14.59}) & 84.51 (\textcolor{blue}{4.28}) & 74.58 & 1.883 & 40.66 \\
\textbf{CU}        & \textbf{0.00} & 74.33 & \textbf{0.00} & 76.23 & 82.57 (\textcolor{blue}{5.27}) & 64.25 (\textcolor{blue}{10.14}) & 85.01 (\textcolor{blue}{3.78}) & 85.07 & 1.451 & 36.64 \\
\midrule
\textbf{EC}        & 0.01 & 74.95 & 0.02 & 76.67 & 82.57 (\textcolor{blue}{5.27}) & 61.49 (\textcolor{blue}{12.90}) & 84.42 (\textcolor{blue}{4.37}) & 78.24 & \textbf{0.477} & \textbf{62.51} \\
\bottomrule
\end{tabular}
}
\caption{Comparison of unlearning methods on ImageNet-1K (\textbf{Swin-Tiny}, random 100 class forgetting). Table notation follows Table~\ref{tab:table_top100}.}
\label{tab:swin_tiny_random100}
\end{table*}

\subsection{EC Combined with Other Unlearning Baselines.}
\label{sec:plugin}

To verify that EC can serve as a model-agnostic plug-in module, we apply EC to other representation-based unlearning baselines and evaluate their EC-augmented variants. Table~\ref{tab:plugin} presents the comparison between baseline methods and their EC-augmented counterparts (denoted +EC). Integrating EC consistently improves representation-based metrics (CKA and $|\text{IDI}|$) while maintaining or slightly improving utility metrics. For instance, DUCK (+EC) improves H-Mean from 44.65 to 50.12, demonstrating that EC can be applied as a practical plug-in to strengthen representation-level forgetting across different unlearning methods.

\begin{table*}[t]
\centering
\scriptsize
\resizebox{\textwidth}{!}{%

\begin{tabular}{l cccc ccc cc c}
\toprule
& \multicolumn{4}{c}{\textbf{ImageNet-1K}} & \multicolumn{3}{c}{\textbf{k-NN}} & & & \\
\cmidrule(lr){2-5} \cmidrule(lr){6-8}
\textbf{Method} & \textbf{FA}$\downarrow$ & \textbf{RA}$\uparrow$ & \textbf{TFA}$\downarrow$ & \textbf{TRA}$\uparrow$ &
\textbf{Office-Home} (\textcolor{blue}{$\downarrow$}) & \textbf{CUB} (\textcolor{blue}{$\downarrow$}) & \textbf{DomainNet-126} (\textcolor{blue}{$\downarrow$}) &
\textbf{CKA}$\downarrow$ & \textbf{$|$IDI$|$} $\downarrow$ & \textbf{H-Mean}$\uparrow$ \\
\midrule
\textbf{Original}     & 78.98 & 80.01 & 76.10 & 76.47 & 80.28 (\textcolor{blue}{1.95}) & 43.00 (\textcolor{blue}{2.08}) & 72.67 (\textcolor{blue}{10.44}) & 100 & 1.000 & -- \\
\textbf{Retrained}    & 0.00 & 84.47 & 0.00 & 77.62 & 78.33 (\textcolor{blue}{0.00}) & 40.92 (\textcolor{blue}{0.00}) & 83.11 (\textcolor{blue}{0.00}) & 86.19 & 0.000 & -- \\
\midrule
\textbf{CU}           & \textbf{0.00} & \textbf{75.83} & \textbf{0.00} & \textbf{75.49} & 80.62 (\textcolor{blue}{2.29}) & 50.42 (\textcolor{blue}{9.50}) & \textbf{83.73 (\textcolor{blue}{0.62})} & 69.52 & 0.403 & 70.68 \\
\textbf{EC}     & \textbf{0.00} & 72.63 & \textbf{0.00} & 73.84 & \textbf{76.83 (\textcolor{blue}{1.50})} & \textbf{44.95 (\textcolor{blue}{4.03})} & 80.88 (\textcolor{blue}{2.23}) & \textbf{38.68} & \textbf{0.051} & \textbf{85.75} \\
\midrule
\textbf{DUCK}         & 0.04 & 71.21 & 0.02 & 72.34 & \textbf{78.67 (\textcolor{blue}{0.34})} & 37.96 (\textcolor{blue}{2.96}) & 81.49 (\textcolor{blue}{1.62}) & 90.15 & 0.538 & 44.65 \\
\textbf{DUCK (+EC)}   & \textbf{0.00} & \textbf{72.02} & \textbf{0.00} & \textbf{73.00} & 80.96 (\textcolor{blue}{2.63}) & \textbf{41.77 (\textcolor{blue}{0.85})} & \textbf{82.65 (\textcolor{blue}{0.46})} & \textbf{87.67} & \textbf{0.511} & \textbf{50.12} \\
\midrule
\textbf{COLA}         & \textbf{0.00} & \textbf{72.57} & \textbf{0.00} & \textbf{73.77} & 78.90 (\textcolor{blue}{0.57}) & \textbf{38.63 (\textcolor{blue}{2.29})} & 81.01 (\textcolor{blue}{2.10}) & 89.28 & 0.867 & 36.54 \\
\textbf{COLA (+EC)}   & \textbf{0.00} & 72.43 & \textbf{0.00} & 73.75 & \textbf{77.98 (\textcolor{blue}{0.35})} & 38.30 (\textcolor{blue}{2.62}) & \textbf{81.60 (\textcolor{blue}{1.51})} & \textbf{89.15} & \textbf{0.846} & \textbf{38.33} \\
\bottomrule
\end{tabular}
}
\caption{Full table of Comparison of baseline methods with and without EC on ImageNet-1K (ResNet-50, random 100 class forgetting). (+EC) denotes the application of EC's multi-layer supervision to each baseline. The better values in each comparison is shown in \textbf{bold}.}
\label{tab:plugin}
\end{table*}

\subsection{Hyperparameter Sensitivity Analysis}
\label{sec:hyperparameter_sensitivity}

Table~\ref{tab:ec_hparam_ablation} presents an ablation study on the key hyperparameters of EC: layer-wise loss weights $w_{1\text{-}4}$ and loss coefficients $(\lambda_{\text{CU}}, \lambda_{\text{CE}})$. Several observations emerge: (1) Assigning larger weights to deeper layers ($w_{1\text{-}4} = (0.2, 0.4, 0.8, 1.0)$) yields the best H-Mean (85.75), consistent with the intuition that deeper layers encode more class-discriminative features. (2) Uniform weights ($w_{1\text{-}4} = (0.5, 0.5, 0.5, 0.5)$) or inverted weights ($w_{1\text{-}4} = (1.0, 0.8, 0.4, 0.2)$) result in lower H-Mean scores. (3) Balancing $\lambda_{\text{CU}}$ and $\lambda_{\text{CE}}$ at $(1.5, 1.5)$ provides a good trade-off between forgetting strength and utility preservation. Based on these findings, we adopt $w_{1\text{-}4} = (0.2, 0.4, 0.8, 1.0)$ and $(\lambda_{\text{CU}}, \lambda_{\text{CE}}) = (1.5, 1.5)$ as the default configuration for EC throughout our experiments.

\begin{table*}[t]
\centering
\scriptsize
\resizebox{\textwidth}{!}{
\begin{tabular}{l c cccc ccc cc c}
\toprule
& & \multicolumn{4}{c}{\textbf{ImageNet-1K}} & \multicolumn{3}{c}{\textbf{k-NN}} & & & \\
\cmidrule(lr){3-6} \cmidrule(lr){7-9}
\textbf{${w_{1\text{-}4}}$} &
\textbf{$\mathbf{(\lambda_{\text{CU}}, \lambda_{\text{CE}})}$} &
\textbf{FA}$\downarrow$ & \textbf{RA}$\uparrow$ & \textbf{TFA}$\downarrow$ & \textbf{TRA}$\uparrow$ &
\textbf{Office-Home} (\textcolor{blue}{$\downarrow$}) & \textbf{CUB} (\textcolor{blue}{$\downarrow$}) & \textbf{DomainNet-126} (\textcolor{blue}{$\downarrow$}) &
\textbf{CKA}$\downarrow$ & \textbf{$|$IDI$|$}$\downarrow$ & \textbf{H-Mean}$\uparrow$ \\
\midrule
(0.2, 0.4, 0.8, 1.0) & (1.5, 1.5) &
\textbf{0.00} & 72.63 & \textbf{0.00} & 73.84 &
\textbf{76.83 (\textcolor{blue}{1.50})} & 44.95 (\textcolor{blue}{4.03}) & 80.88 (\textcolor{blue}{2.23}) &
38.68 & \textbf{0.051} & \textbf{85.75} \\
(0.5, 0.5, 0.5, 0.5) & (1.5, 1.5) &
\textbf{0.00} & 71.03 & \textbf{0.00} & 72.49 &
65.60 (\textcolor{blue}{12.73}) & 46.73 (\textcolor{blue}{5.81}) & 81.54 (\textcolor{blue}{1.57}) &
43.20 & 0.222 & 81.40 \\
(1.0, 0.8, 0.4, 0.2) & (1.5, 1.5) &
0.21 & 73.49 & 0.30 & \textbf{74.29} &
68.35 (\textcolor{blue}{9.98}) & 49.32 (\textcolor{blue}{8.40}) & \textbf{82.84 (\textcolor{blue}{0.27})} &
\textbf{33.49} & 0.328 & 82.51 \\
(0.1, 0.2, 0.6, 1.0) & (1.5, 1.5) &
\textbf{0.00} & 71.03 & \textbf{0.00} & 72.49 &
64.68 (\textcolor{blue}{13.65}) & 48.69 (\textcolor{blue}{7.77}) & 80.71 (\textcolor{blue}{2.40}) &
39.92 & 0.074 & 83.35 \\
(0.0, 0.0, 1.0, 1.0) & (1.5, 1.5) &
\textbf{0.00} & 69.66 & 0.02 & 71.46 &
64.79 (\textcolor{blue}{13.54}) & 46.40 (\textcolor{blue}{5.48}) & 80.08 (\textcolor{blue}{3.03}) &
35.73 & 0.100 & 83.74 \\
(0.2, 0.4, 0.8, 1.0) & (2.0, 1.0) &
\textbf{0.00} & 69.61 & \textbf{0.00} & 71.97 &
61.47 (\textcolor{blue}{16.86}) & \textbf{44.32 (\textcolor{blue}{3.40})} & 78.71 (\textcolor{blue}{4.40}) &
53.45 & 0.155 & 78.66 \\
(0.2, 0.4, 0.8, 1.0) & (1.0, 2.0) &
\textbf{0.00} & \textbf{73.67} & \textbf{0.00} & 74.00 &
70.76 (\textcolor{blue}{7.57}) & 47.24 (\textcolor{blue}{6.32}) & 82.81 (\textcolor{blue}{0.30}) &
53.84 & 0.139 & 80.44 \\
\bottomrule
\end{tabular}
}
\caption{Ablation of EC hyperparameters on ImageNet-1K (ResNet-50, random 100 class forgetting): layer-wise loss weights $w_{1\text{-}4}$ and loss coefficients $(\lambda_{\text{CU}}, \lambda_{\text{CE}})$. $|$IDI$|$ denotes the absolute value of IDI. For k-NN results, \textcolor{blue}{blue numbers} indicate the absolute gap relative to the Retrained baseline. \textbf{Bold} indicates the best result among the listed EC settings (for k-NN, the smallest absolute gap with retrained model).}
\label{tab:ec_hparam_ablation}
\end{table*}

\section{Experimental Details}
\label{sec:experimental_details}

\subsection{Experimental Setup}
\label{sec:setup}
All unlearning experiments were conducted on a single NVIDIA RTX 4090 GPU (24GB VRAM) with an Intel Xeon Gold 6426Y CPU. We used Ubuntu 22.04.3 LTS, CUDA 11.7, PyTorch 2.0.1, and torchvision 0.15.2. Unless otherwise stated, automatic mixed precision (AMP) was enabled for all experiments.


\subsection{Pretraining Details for Original and Retrained Models}
\label{sec:pretraining}

We follow the pretraining protocol of \citeauthor{kim2025arewe}~\shortcite{kim2025arewe} whenever applicable. All pretraining and fine-tuning runs were conducted using 4$\times$ RTX 4090 GPUs (24GB each). Table~\ref{tab:pretraining} summarizes the configurations.

For ResNet-50 on ImageNet-1K, we train both the Original and Retrained (retain-only) models from scratch under the same optimization recipe using SGD with momentum 0.9 and a step learning rate scheduler. For CIFAR-100, we use the same recipe except for epoch and batch size. For Swin-Tiny, we initialize with ImageNet-21K pretrained weights following \citeauthor{jeon2024idi}~\shortcite{jeon2024idi} and fine-tune on ImageNet-1K for 30 epochs using AdamW with cosine annealing.

\begin{table*}[t]
\centering
\scriptsize
\resizebox{0.9\textwidth}{!}{%
\setlength{\tabcolsep}{7pt}
\renewcommand{\arraystretch}{1.05}
\begin{tabular}{lccc}
\toprule
\multicolumn{4}{c}{Pretraining / Fine-tuning Recipes} \\
\cmidrule(lr){2-4}
Settings & ResNet-50 (ImageNet-1K) & Swin-Tiny (ImageNet-1K) & ResNet-50 (CIFAR-100) \\
\midrule
Initialization& Scratch& ImageNet-21K pretrained& Scratch \\
Target data& ImageNet-1K& ImageNet-1K& CIFAR-100 \\
Epochs& 182& 30& 150 \\
Batch Size& 256& 256& 128 \\
Learning Rate& 0.1& 0.02& 0.1 \\
Optimizer& SGD & AdamW & SGD \\
Momentum& 0.9 & 0.9 & 0.9 \\
Scheduler& Step& CosineAnnealing& Step \\
\bottomrule
\end{tabular}
}
\caption{Backbone pretraining/fine-tuning configurations.
For all runs, the same recipe is used to train both the Original model and the Retrained (retain-only) baseline from scratch. All runs use 4$\times$ RTX 4090 (24GB).}
\label{tab:pretraining}
\end{table*}

\subsection{EC Architecture and Pre-training of EC Modules}
\label{sec:ec_architecture}

We attach EC modules after intermediate stages of the backbone. For ResNet-50, EC modules are attached after Stages 1, 2, and 3, but not after Stage 4. At Stage 4, the contrastive unlearning loss $\mathcal{L}_{\text{CU}}^4$ is computed on features extracted after the global average pooling layer, and the cross-entropy loss $\mathcal{L}_{\text{CE}}^4$ is computed on the logits from the final classifier. For Swin-Tiny, we follow the same design: EC modules are attached after Stages 1, 2, and 3, with no additional module after Stage 4.

Each EC module consists of a sequence of convolutional (Conv) blocks followed by a fully connected (FC) classifier. For a backbone with $N$ stages, the EC module attached after Stage $k$ contains $(N-k)$ Conv blocks. The detailed architecture of each Conv block follows \citeauthor{zhang2022contrastivedeepsup}~\shortcite{zhang2022contrastivedeepsup}.

Before unlearning, we pre-train the Conv blocks of each EC module on the full dataset $\mathcal{D}$ using Supervised Contrastive Learning (SupCon)~\cite{khosla2020supervised}, while keeping the backbone and final classifier frozen. For both ResNet-50 and Swin-Tiny, we pre-train for 80 epochs with a batch size of 1024 and a learning rate of $2 \times 10^{-2}$, using 4$\times$ RTX 4090 GPUs (24GB each). The FC classifiers within each EC module are randomly initialized and remain untrained during this phase; they are learned jointly with the backbone through the cross-entropy loss $\mathcal{L}_{\text{CE}}^l$ during the unlearning stage.

\subsection{Hyperparameter Details}
\label{sec:baselines}

We implement all baseline methods following their original papers and official codebases when available. Table~\ref{tab:hyperparams} lists the detailed hyperparameters for each method. To ensure a fair comparison, we conducted hyperparameter search for each baseline under our evaluation protocol, selecting settings that maximize overall performance (measured by H-Mean). For methods without official implementations (e.g., DELETE), we re-implemented them based on the descriptions in the original papers. All baselines use the same pre-trained Original model as the starting point for unlearning.

\begin{table*}[t]
\centering
\scriptsize
\resizebox{\textwidth}{!}{
\begin{tabular}{l l p{0.45\textwidth} c}
\toprule
\textbf{Methods} & \textbf{Hyperparameter} & \textbf{Description of hyperparameters} & \textbf{Values} \\
\midrule
\multirow{3}{*}{\textbf{DUCK}~\cite{cotogni2023duck}}
& $\lambda_{\text{fgt}}$ & Weight for the forgetting loss (distance to incorrect centroids) & 1.5 \\
& $\lambda_{\text{ret}}$ & Weight for the retaining loss (distance to correct centroids) & 1.5 \\
& $\tau$ & Temperature scaling parameter for distance-based softmax & 2.0 \\
\midrule
\multirow{3}{*}{\textbf{SCAR}~\cite{bonato2024retain}}
& $\lambda_1$ & Weight for the metric unlearning loss & 5.0 \\
& $\lambda_2$ & Weight for the distillation loss on OOD data & 5.0 \\
& $\gamma_1, \gamma_2$ & Covariance shrinkage coefficients for distribution estimation & 3.0, 3.0 \\
\midrule
\multirow{3}{*}{\textbf{SCRUB}~\cite{kurmanji2023towards}}
& $\alpha$ & Weight for knowledge distillation loss on the retain set & 1.0 \\
& $\gamma$ & Weight for cross-entropy loss on the retain set & 1.0 \\
& m\_steps & Number of optimization steps for the maximization phase & 200 \\
\midrule
\textbf{SalUn}~\cite{fan2024salun}
& $pt$ & Sparsity ratio threshold for gradient-based weight masking & 0.5 \\
\midrule
\multirow{2}{*}{\textbf{DELETE}~\cite{zhou2025decoupled}}
& $\alpha$ & Scaling factor for the target probability of the forget class & 0 \\
& $\tau$ & Temperature parameter for mask distillation & 1.0 \\
\midrule
\multirow{4}{*}{\textbf{CU}~\cite{zhang2024contrastive}}
& $\tau$ & Temperature parameter for contrastive loss & 0.07 \\
& $\omega$ & Number of negative batches sampled per anchor & 2 \\
& $\lambda_{\text{UL}}$ & Weight for the contrastive unlearning loss term & 1.0 \\
& $\lambda_{\text{CE}}$ & Weight for the cross-entropy retention loss term & 1.0 \\
\midrule
\multirow{5}{*}{\textbf{EC (Ours)}}
& $\tau$ & Temperature parameter for contrastive unlearning & 0.07 \\
& $\omega$ & Number of retain set batches sampled per forget set anchor & 2 \\
& $\lambda_{\text{CU}}$ & Weight for the multi-layer contrastive unlearning loss & 1.5 \\
& $\lambda_{\text{CE}}$ & Weight for the cross entropy (retention) loss & 1.5 \\
& $w_{1-4}$ & Layer-wise weights for EC modules & 0.2, 0.4, 0.8, 1.0 \\
\bottomrule
\end{tabular}
}
\caption{Detailed hyperparameters for the baseline methods and our proposed method (EC). For each method, hyperparameters not explicitly listed follow the default settings in the original paper/implementation. Reported values were selected by repeated hyperparameter search (multiple runs) to find settings that are well-suited to our experimental protocol, ensuring a fair comparison across methods.}
\label{tab:hyperparams}
\end{table*}

\subsection{Unlearning Details}
\label{sec:unlearning_detail}

Table~\ref{tab:unlearning_detail} summarizes the learning rate, number of epochs, and batch size used for each unlearning method. These settings were determined through hyperparameter search to achieve optimal performance under our evaluation protocol. For COLA, we report epochs for the Collapse and Align stages separately (1 and 2 epochs, respectively). All methods use the Adam or SGD optimizer depending on the original implementation, with the learning rates specified in the table.

\begin{table*}[t]
\centering
\small
\resizebox{0.5\textwidth}{!}{
\begin{tabular}{l c c c}
\toprule
\textbf{Method} & \textbf{Learning rate} & \textbf{Epochs} & \textbf{Batch size} \\
\midrule
\textbf{PL}                     & $5\times10^{-5}$ & 20   & 256 \\
\textbf{DUCK}                   & $1\times10^{-3}$ & 14   & 64  \\
\textbf{SCAR}                   & $5\times10^{-4}$ & 38   & 128 \\
\textbf{SCRUB}                  & $5\times10^{-5}$ & 21   & 128 \\
\textbf{SalUn}                  & $5\times10^{-6}$ & 10   & 256 \\
\textbf{RL}                     & $5\times10^{-6}$ & 13   & 256 \\
\textbf{DELETE}                 & $5\times10^{-4}$ & 40   & 128 \\
\textbf{COLA}                   & $5\times10^{-5}$ & 1, 2 & 256 \\
\textbf{CU}                     & $2\times10^{-4}$ & 20   & 128 \\
\textbf{EC}                     & $2\times10^{-4}$ & 20   & 128 \\
\bottomrule
\end{tabular}
}
\caption{Unlearning settings for each method (learning rate, number of epochs, and batch size). For COLA, ``1, 2'' denote the epochs for the Collapse and Align stages, respectively. The reported settings were selected via repeated experimental runs and hyperparameter search, choosing the configuration that yielded the best overall performance under our evaluation protocol for a fair comparison.}
\label{tab:unlearning_detail}
\end{table*}

\subsection{Evaluation Implementation Details}
\label{sec:evaluation}
\paragraph{CKA Computation.} We compute linear Centered Kernel Alignment (CKA)~\cite{kornblith2019similarity} between the original model $f_o$ and the unlearned model $f_u$ using features extracted from the test forget set $\mathcal{D}_f^{\text{te}}$. For layer-wise CKA analysis, we extract features from intermediate bottleneck blocks within Layer 4 of ResNet-50 (denoted Layer 4.0, 4.1, and 4.2). Notably, these features are extracted directly from the backbone and passed through global average pooling, independent of the EC modules.


\paragraph{IDI Computation.} We follow the original implementation of \citeauthor{jeon2024idi}~\cite{jeon2024idi} to compute the Information Difference Index (IDI). IDI measures the residual mutual information between intermediate features and forget labels by comparing the unlearned model to a retrained baseline. We compute IDI using features from the last three bottleneck blocks of Layer 4 in ResNet-50. We use the following hyperparameters: batch size of 512, 5 training epochs with a maximum of 800 steps per epoch, output feature dimension $d = 128$ for the critic functions, learning rate of $3 \times 10^{-5}$ for $f_{\nu_\ell}$, and learning rate of $5 \times 10^{-4}$ for $g_{\eta_\ell}$.

\paragraph{k-NN Downstream Evaluation.} Following \citeauthor{kim2025arewe}~\shortcite{kim2025arewe}, we freeze the backbone of each unlearned model and train a $k$-nearest neighbor classifier ($k=5$) on features extracted from three downstream datasets: Office-Home~\cite{venkateswara2017}, CUB-200-2011~\cite{catherineundefined}, and DomainNet-126~\cite{peng2018}. We report the accuracy and the absolute gap relative to the Retrained baseline.

\paragraph{H-Mean Computation.}
We normalize all metrics to a common scale $[0,100]$ where larger is better, and compute the harmonic mean over nine scores:
\begin{equation}
\mathrm{H\text{-}Mean}
= \frac{9}{\sum\frac{1}{s}} .
\label{eq:hmean}
\end{equation}

\noindent
The normalized scores are:
\begin{equation}
\begin{aligned}
s_{\mathrm{FA}}  &= 100 - \mathrm{FA}, \\
s_{\mathrm{TFA}} &= 100 - \mathrm{TFA}, \\
s_{\mathrm{CKA}} &= 100 - \mathrm{CKA}, \\
s_{\mathrm{RA}}  &= \mathrm{RA}, \\
s_{\mathrm{TRA}} &= \mathrm{TRA}, \\
s_{\mathrm{kNN}} &= 100 - \lvert \mathrm{gap} \rvert, \\
s_{\mathrm{IDI}} &= 100 \cdot \operatorname{clip}\ \!\bigl(1-\lvert \mathrm{IDI}\rvert,\ 0.1,\ 1\bigr).
\\
\end{aligned}
\label{eq:normalized_scores}
\end{equation}

\noindent
Here $\operatorname{clip}(x,a,b)=\min(\max(x,a),b)$.
For $\lvert \mathrm{IDI}\rvert$, we compute $1-\lvert \mathrm{IDI}\rvert$ and clip it to $[0.1,1]$ before scaling by 100 to improve numerical stability in \eqref{eq:hmean}.


\end{document}